\documentclass{article}

\PassOptionsToPackage{numbers, compress}{natbib}

\usepackage[preprint]{neurips_2025}

\usepackage[utf8]{inputenc} 
\usepackage[T1]{fontenc}    
\usepackage{hyperref}       
\usepackage{url}            
\usepackage{booktabs}       
\usepackage{amsfonts}       
\usepackage{nicefrac}       
\usepackage{microtype}      
\usepackage{xcolor}         
\usepackage{amsmath}
\usepackage{graphicx} 
\usepackage{svg}
\usepackage{wrapfig}
\usepackage{multirow}

\newcommand{\LCE}{\mathcal{L}_{\text{CE}}}
\newcommand{\Pmodel}{P_{\text{model}}}
\newcommand{\Ptrue}{P_{\text{true}}}

\newcommand{\DKL}{D_{\text{KL}}}

\newcommand{\Ieff}{I_{\text{eff}}}

\newcommand{\E}{\mathbb{E}}

\newcommand{\uparr}{\textcolor{red}{$\uparrow$}}
\newcommand{\downarr}{\textcolor{green}{$\downarrow$}}

\title{TinyAlign: Boosting Lightweight Vision-Language Models by Mitigating Modal Alignment Bottlenecks}

\author{%
Yuanze Hu\(^1\)
\And
Zhaoxin Fan\(^{1,2}\)\thanks{Corresponding author}
\And
Xinyu Wang\(^1\)
\And
Gen Li\(^1\)
\And
Ye Qiu\(^1\)
\And
Zhichao Yang\(^1\)
\And
Wenjun Wu\(^{1,2}\)
\And
Kejian Wu\(^3\)
\And
Yifan Sun\(^4\)
\And
Xiaotie Deng\(^5\)
\And
Jin Dong\(^6\)
\\
\(^1\) Beijing Advanced Innovation Center for Future Blockchain and Privacy Computing,\\ Beihang University 
\(^2\) Hangzhou International Innovation Institute, Beihang University \\
\(^3\) Xreal ,
\(^4\) Renmin University ,
\(^5\) Peking University \\
\(^6\) Beijing Academy of Blockchain and Edge Computing (BABEC)
}

\begin{document}

\maketitle

\begin{abstract}
Lightweight Vision-Language Models (VLMs) are indispensable for resource-constrained applications. The prevailing approach to aligning vision and language models involves freezing both the vision encoder and the language model while training small connector modules. However, this strategy heavily depends on the intrinsic capabilities of the language model, which can be suboptimal for lightweight models with limited representational capacity. In this work, we investigate this alignment bottleneck through the lens of mutual information, demonstrating that the constrained capacity of the language model inherently limits the Effective Mutual Information (EMI) between multimodal inputs and outputs, thereby compromising alignment quality. To address this challenge, we propose TinyAlign, a novel framework inspired by Retrieval-Augmented Generation, which strategically retrieves relevant context from a memory bank to enrich multimodal inputs and enhance their alignment. Extensive empirical evaluations reveal that TinyAlign significantly reduces training loss, accelerates convergence, and enhances task performance. Remarkably, it allows models to achieve baseline-level performance with only 40\% of the fine-tuning data, highlighting exceptional data efficiency. Our work thus offers a practical pathway for developing more capable lightweight VLMs while introducing a fresh theoretical lens to better understand and address alignment bottlenecks in constrained multimodal systems.
\end{abstract}

\section{Introduction}
\label{sec:introduction}
The rapid advancements in Large Language Models (LLMs) have catalyzed the development of Vision-Language Models (VLMs), enabling models to excel in complex multimodal reasoning and understanding tasks. Prominent models such as Gemini 2.5 Pro~\cite{gemini2.5pro2024}, GPT-4V~\cite{gpt4v2023}, Qwen2.5-VL 72B~\cite{bai2025qwen25vltechnicalreport}, and PaLI-X~\cite{chen2023palixscalingmultilingualvision} have showcased remarkable performance across various benchmarks, setting new standards for multimodal intelligence. However, these models typically involve billions of parameters, resulting in significant computational and storage demands. Such massive requirements make these models impractical for resource-constrained scenarios, such as edge devices or applications with limited computational budgets. This growing demand for efficiency has turned lightweight VLMs into a critical area of research, as they aim to retain strong multimodal capabilities while drastically reducing computational costs and memory footprints, thereby enabling broader applicability.

To achieve this balance between performance and efficiency, most  VLMs adopt a modular design where pre-trained vision encoders and language models are frozen, and a small "connector" module is trained to align the two modalities. This approach, also seen in lightweight models such as MiniGPT-4~\cite{zhu2023minigpt4enhancingvisionlanguageunderstanding}, BLIP-2~\cite{li2023blip2bootstrappinglanguageimagepretraining}, and Visual Instruction Tuning~\cite{liu2023visualinstructiontuning}, is computationally efficient and leverages the strong representational power of pre-trained components. However, while effective, this implicit alignment paradigm might struggles in the context of lightweight VLMs. The limited representational capacity of smaller LLMs significantly constrains their ability to process and align multimodal information, leading to subpar performance on complex tasks. This misalignment becomes a critical bottleneck for lightweight VLMs, as it prevents these models from fully realizing their potential.

To address this issue, we conduct a theoretical analysis of the alignment bottleneck in lightweight VLMs from the perspective of information theory\cite{zhu2024informationbottleneckperspectiveeffective,liu2025pointwisemutualinformationperformance}. Specifically,  we introduce the concept of Effective Mutual Information (EMI) to quantify the amount of information a model can practically leverage given its capacity constraints. Our analysis reveals that freezing the parameters of a lightweight language model inherently limits the EMI, creating a bottleneck that restricts the model’s ability to align multimodal inputs effectively. This limitation manifests in suboptimal learning dynamics when using standard training objectives, such as cross-entropy loss, and highlights the need for new strategies to enhance the effective flow of information during alignment.

Motivated by this insight, we propose TinyAlign, a novel pre-training and fine-tuning framework explicitly designed to overcome this alignment bottleneck. Inspired by Retrieval-Augmented Generation (RAGs) \cite{lewis2021retrievalaugmentedgenerationknowledgeintensivenlp, guu2020realmretrievalaugmentedlanguagemodel, hu2023revealretrievalaugmentedvisuallanguagepretraining}
, TinyAlign introduces a memory bank constructed from multimodal training instances within the dataset, setting it apart from traditional approaches that depend on external knowledge sources. During training, the framework retrieves contextually relevant representations from the memory bank and augments the original visual inputs with enriched multimodal context. By increasing the effective information content available to the model, TinyAlign reduces the inherent learning difficulty posed by the limited capacity of lightweight language models. This design not only enhances alignment but also optimizes the utilization of available training data, addressing the core challenges faced by lightweight VLMs.

We validate the effectiveness of TinyAlign through extensive experiments across a diverse set of lightweight architectures, including Vicuna~\cite{vicuna2023}, Phi-2~\cite{gunasekar2023textbooksneed}, TinyLLaMA~\cite{zhang2024tinyllamaopensourcesmalllanguage}, and Qwen2~\cite{yang2024qwen2technicalreport}, as well as vision encoders like SigLIP~\cite{zhai2023sigmoidlosslanguageimage} and CLIP~\cite{radford2021learningtransferablevisualmodels}. Our results demonstrate that TinyAlign significantly accelerates convergence, reduces alignment losses (Fig.~\ref{fig:tinyalign_architecture}(a)), and produces more compact and meaningful feature representations (Fig.~\ref{fig:tinyalign_architecture}(b)). Furthermore, TinyAlign exhibits exceptional data efficiency, achieving baseline-level performance with only 40\% of the fine-tuning data.  Our contribution can be summarized as:

\begin{itemize}
    \item We identify a fundamental alignment bottleneck in lightweight VLMs, theoretically analyzing its root cause using principles from information theory. Our analysis reveals that the limited capacity of lightweight LLMs constrains the effective information flow during multimodal alignment.    
    \item We propose TinyAlign, a novel framework inspired by Retrieval-Augmented Generation, which enriches multimodal inputs by retrieving relevant context from a memory bank. This approach enhances the effective information content accessible to the model, addressing the alignment bottleneck.
    \item We conduct extensive experiments to validate the effectiveness of TinyAlign, demonstrating notable improvements in convergence speed, alignment quality, downstream task performance, and data efficiency compared to conventional alignment methods.
\end{itemize}

\section{Related Work}
\label{sec:related_work}
\paragraph{The LLM-Centric Paradigm in VLMs.}
The LLM-Centric Paradigm has become a dominant framework in Vision-Language Models, leveraging pre-trained Large Language Models  as the core for cross-modal understanding\cite{yang2024qwen2technicalreport,bai2025qwen25vltechnicalreport,lu2024deepseekvlrealworldvisionlanguageunderstanding,li2023blip2bootstrappinglanguageimagepretraining,chen2023minigptv2largelanguagemodel,chen2024internvlscalingvisionfoundation}. This approach typically freezes the parameters of both the vision encoder and the LLM~\cite{li2023blip2bootstrappinglanguageimagepretraining}, while training a lightweight connector module to bridge vision and language~\cite{yang2024qwen2technicalreport}. This implicit alignment strategy has achieved notable empirical success, as seen in models like DeepSeek-VL~\cite{lu2024deepseekvlrealworldvisionlanguageunderstanding} and Qwen2.5-VL~\cite{bai2025qwen25vltechnicalreport}. However, the theoretical mechanisms enabling effective cross-modal harmonization remain underexplored, with most research focusing on empirical results rather than systematic analysis. To fill this gap, our work conducts one of the first in-depth theoretical investigations into this paradigm, uncovering its principles and limitations to better understand cross-modal alignment.
\paragraph{Advancements in Lightweight VLMs.}
Recent efforts to create lightweight Vision-Language Models (VLMs) have explored various avenues for enhancing efficiency and performance\cite{yuan2024tinygptvefficientmultimodallarge,zhou2024tinyllavaframeworksmallscalelarge,yao2024minicpmvgpt4vlevelmllm,marafioti2025smolvlmredefiningsmallefficient,steiner2024paligemma2familyversatile}.
EfficientVLM~\cite{wang2022efficientvlm} introduces a distill-then-prune framework with modal-adaptive pruning to compress large VLMs effectively.
TinyLLaVA~\cite{zhou2024tinyllavaframeworksmallscalelarge} explores optimal pairings of language models, vision encoders, and connectors for small-scale VLMs, while MobileVLM and its successor v2 emphasize architectural innovations, high-quality data, and advanced training strategies~\cite{chu2023mobilevlmfaststrong,chu2024mobilevlmv2fasterstronger}, while SmolVLM~\cite{marafioti2025smolvlmredefiningsmallefficient} further explores tokenization strategies.MiniCPM-V~\cite{yao2024minicpmvgpt4vlevelmllm} presents a series of efficient Multimodal Large Language Models (MLLMs) designed for on-device deployment, achieved by integrating advanced techniques in architecture, pre-training, and alignment.
However, these approaches primarily focus on component optimization, model compression, advanced training strategies, or designing for edge deployment, and seldom question whether the widely adopted implicit alignment paradigm is fundamentally suitable for models with limited capacity.
In contrast, we provide a principled analysis demonstrating that this paradigm intrinsically induces higher alignment loss for smaller models, thereby limiting their potential for robust visual understanding and cross-modal alignment.
\paragraph{Retrieval-Augmented Models.}
Retrieval-Augmented Generation  enhances factual accuracy in NLP by integrating external knowledge retrieval with parametric models~\cite{lewis2021retrievalaugmentedgenerationknowledgeintensivenlp}. Techniques like unsupervised retriever pre-training enable efficient access to large-scale documents during training and inference~\cite{guu2020realmretrievalaugmentedlanguagemodel}. RAG now extends to multimodal tasks, with MM-REACT combining language models and vision experts for complex reasoning~\cite{yang2023mmreactpromptingchatgptmultimodal}, and Re-ViLM reducing parameters by storing knowledge externally for image-to-text generation~\cite{yang2023revilmretrievalaugmentedvisuallanguage}. Frameworks like RAVEN and MuRAG apply retrieval for multitask learning and open-domain question answering~\cite{rao2024ravenmultitaskretrievalaugmented, chen2022muragmultimodalretrievalaugmentedgenerator}, while models like REVEAL unify memory, retrieval, and generation across diverse sources~\cite{hu2023revealretrievalaugmentedvisuallanguagepretraining}. These approaches rely on large external memory banks and retriever modules to broaden knowledge for reasoning-heavy tasks~\cite{caffagni2024wikillavahierarchicalretrievalaugmentedgeneration,hu2023revealretrievalaugmentedvisuallanguagepretraining,rao2024ravenmultitaskretrievalaugmented}. In contrast, TinyAlign addresses the EMI bottleneck in lightweight VLMs by constructing a memory bank from multimodal \emph{training instances}. During pre-training and fine-tuning, TinyAlign retrieves relevant representations to augment visual input, increasing mutual information between inputs and outputs and overcoming the limitations of compact models.

\section{Theoretical Framework: LLM-Centric Alignment and Its Limitations}
\label{sec:theoretical_framework}
\subsection{Cross-Entropy Loss in LLM-Centric VLM Pre-training}
We begin with an analysis of the commonly used LLM-centric paradigm for Vision-Language Models, where the objective is to align visual information with a pre-trained LLM. In this setup, the visual input is denoted as \(X_V\), the accompanying textual instruction as \(X_I\), and the target output, often a textual description or answer, as \(L\). 

The processing pipeline in this paradigm is structured as follows:  
\begin{enumerate}
\item A frozen Vision Encoder (e.g., a Vision Transformer, ViT) with parameters \(\theta_{\text{ViT}}\) processes the visual input \(X_V\) to extract visual features: \(\, Z_V = \text{ViT}(X_V; \theta_{\text{ViT}})\).

\item A trainable Connector module, parameterized by \(\theta_C^*\), transforms the visual features \(Z_V\) into embeddings \(H_V\) compatible with the LLM's input space: \(\, H_V = \text{Connector}(Z_V; \theta_C^*)\).

\item Simultaneously, the textual instruction \(X_I\) is processed by the frozen LLM’s input embedding layer (parameterized by \(\theta_{\text{LLM}}\)), generating instruction embeddings: \(\, H_I = \text{LLM}_{\text{embed}}(X_I; \theta_{\text{LLM}})\).

\item The visual embeddings \(H_V\) and instruction embeddings \(H_I\) are concatenated to form the joint input: \(\, H_{\text{in}} = [H_V, H_I]\).

\item Finally, the frozen LLM produces an output distribution over the target \(L\): \(\, \Pmodel(L | H_{\text{in}}; \theta_{\text{LLM}})\).
\end{enumerate}

\subsection{The LLM Bottleneck: Irreducible Error and Effective Mutual Information}
\label{subsec:llm_bottleneck}

As discussed before, we believe the LLM-centric paradigm inherently suffers from a bottleneck due to its reliance on implicit alignment mechanisms. To understand this limitation, we analyze it from an information-theoretic perspective and show how it constrains the system’s ability to fully utilize the information in the inputs.  The starting point of this analysis is the conditional cross-entropy (CE) loss, which measures the discrepancy between the true conditional distribution of the target labels \(L\) and the model’s predictions. This loss can be decomposed as follows:  
\begin{align}
\LCE(\theta_C^* | X_V, X_I) &= H(\Ptrue(L|X_V, X_I)) \nonumber \\
&\quad + \DKL(\Ptrue(L|X_V, X_I) || \Pmodel(L | [H_V(\theta_C^*), H_I]; \theta_{\text{LLM}})) \label{eq:ce_kl_decomposition_vlm}
\end{align}  
Here, \(H(\Ptrue(L|X_V, X_I))\) represents the true conditional entropy of \(L\) given the inputs \((X_V, X_I)\), which is independent of the model parameters and reflects the irreducible uncertainty in the labels. The second term, the KL divergence, measures the gap between the true conditional distribution and the model’s predictions. Thus, minimizing \(\LCE(\theta_C^*)\) is equivalent to minimizing the KL divergence term, as the conditional entropy is fixed.  

At the beginning of training, the Connector parameters \(\theta_C^*\) are randomly initialized, leading to arbitrary and semantically meaningless visual embeddings \(H_V(\theta_C^*)\) from the perspective of the frozen LLM (\(\theta_{\text{LLM}}\)). When the LLM processes these unfamiliar embeddings \(H_V\) alongside \(H_I\), it produces outputs that deviate significantly from the true target \(L\), resulting in a high initial loss. During training, the Connector learns to transform raw visual features \(Z_V\) into embeddings \(H_V\) that the frozen LLM can interpret coherently with \(H_I\) to predict \(L\). This process can be viewed as the Connector "translating" visual information into the LLM’s fixed semantic space. 
However, even with an optimal Connector \(\theta_C^{\text{opt}}\) that produces the best possible \(H_V\), the frozen LLM’s representational capacity is inherently limited. Its ability to integrate and reason over the novel visual embeddings \(H_V\) is constrained by its fixed architecture and pre-trained knowledge. This introduces an irreducible error, quantified as the average KL divergence that cannot be eliminated even with optimal Connector parameters:  
\[
\bar{\epsilon}_{\theta_{\text{LLM}}} = \E_{(X_V, X_I)} \left[ \min_{\theta_C^*} \DKL(\Ptrue(L|X_V,X_I) || \Pmodel(L | [\text{Connector}(Z_V;\theta_C^*), H_I]; \theta_{\text{LLM}})) \right] \ge 0
\]  
This irreducible error, \(\bar{\epsilon}_{\theta_{\text{LLM}}}\), reflects the frozen LLM’s inability to fully align its semantic space with the optimally transformed visual embeddings \(H_V\).  The minimum achievable CE loss for the system, given the frozen parameters \(\theta_{\text{ViT}}\) and \(\theta_{\text{LLM}}\), can therefore be expressed as:  
\begin{align}
\min_{\theta_C^*} \LCE(\theta_C^*) &= H(L|X_V, X_I) + \bar{\epsilon}_{\theta_{\text{LLM}}} \label{eq:min_loss_constrained_vlm}
\end{align}  
Here, \(H(L|X_V, X_I)\) represents the irreducible uncertainty in the target labels, while \(\bar{\epsilon}_{\theta_{\text{LLM}}}\) quantifies the frozen LLM's limitations.  To further analyze this bottleneck, we rewrite the minimum loss using the definition of mutual information \(I(X; Y) = H(Y) - H(Y|X)\). Substituting \(H(L|X_V, X_I)\) into Eq.~\eqref{eq:min_loss_constrained_vlm}, we obtain:  
\begin{align}
\min_{\theta_C^*} \LCE(\theta_C^*) = H(L) - I(X_V, X_I; L) + \bar{\epsilon}_{\theta_{\text{LLM}}} \label{eq:min_loss_constrained_mi_vlm}
\end{align}  
Here, \(H(L)\) is the entropy of the target labels, and \(I(X_V, X_I; L)\) is the true mutual information between the inputs \((X_V, X_I)\) and the labels \(L\).  To capture the system’s actual capability to utilize information, we define an "Effective Mutual Information" (\(\Ieff\)), which represents the mutual information the system can effectively leverage, accounting for the irreducible error \(\bar{\epsilon}_{\theta_{\text{LLM}}}\):  
\begin{align} 
\Ieff(X_V, X_I; L | \theta_{\text{LLM}}, \theta_{\text{ViT}}) = I(X_V, X_I; L) - \bar{\epsilon}_{\theta_{\text{LLM}}} \label{eq:ieff_definition_vlm} 
\end{align}  
Substituting \(\Ieff\) into Eq.~\eqref{eq:min_loss_constrained_mi_vlm}, the minimum achievable loss becomes:  
\begin{align}
\min_{\theta_C^*} \LCE(\theta_C^*) \approx H(L) - \Ieff(X_V, X_I; L | \theta_{\text{LLM}}, \theta_{\text{ViT}}) \label{eq:min_loss_ieff_vlm} 
\end{align}  
This result highlights a critical limitation of the LLM-centric paradigm: the effective mutual information (\(\Ieff\)) is fundamentally constrained by the frozen LLM’s ability to process and integrate visual embeddings. Even with perfect transformation of visual features by the Connector, the irreducible error \(\bar{\epsilon}_{\theta_{\text{LLM}}}\) reduces the system’s ability to fully utilize the true mutual information \(I(X_V, X_I; L)\). It underscores the need for alternative approaches to achieve better cross-modal alignment.

\textbf{Hypothesis on LLM Scale.}
Based on the theoretical analysis above, we hypothesize that a smaller, lightweight LLM (with parameters \(\theta_{\text{LLM,small}}\)) possesses a more restricted representational capacity and is inherently less capable of interpreting and integrating novel semantic inputs \(H_V\) compared to a larger, more powerful LLM (with parameters \(\theta_{\text{LLM,large}}\)). As shown in Fig.~\ref{fig:combined_loss_and_umap}(a), the larger LLM exhibits a low loss even before ever seeing the image, suggesting its superior capacity to handle semantic information. This implies:  
\[
\bar{\epsilon}_{\theta_{\text{LLM,small}}} \ge \bar{\epsilon}_{\theta_{\text{LLM,large}}}
\]  
Consequently, from Eq.~\eqref{eq:ieff_definition_vlm}, the effective mutual information for a system using a smaller LLM will be lower:  
\[
\Ieff(X_V, X_I; L | \theta_{\text{LLM,small}}, \theta_{\text{ViT}}) \le \Ieff(X_V, X_I; L | \theta_{\text{LLM,large}}, \theta_{\text{ViT}})
\]  
And from Eq.~\eqref{eq:min_loss_constrained_vlm}, its minimum achievable CE loss will be higher:  
\[
\min_{\theta_C^*} \LCE(\text{using } \theta_{\text{LLM,small}}) \ge \min_{\theta_C^*} \LCE(\text{using } \theta_{\text{LLM,large}})
\]  
This theoretical observation explains why VLMs built upon smaller frozen LLMs often converge to higher loss values and exhibit poorer performance. The LLM’s intrinsic capacity to process and integrate the optimally transformed visual embeddings \(H_V\) becomes the bottleneck for the entire VLM’s capability.

\section{TinyAlign: Mitigating Modal Alignment Bottleneck of Lightweight VLMs}
\label{sec:tinyalign_methodology}

\begin{figure}[htbp]
  \centering
  \includegraphics[width=\textwidth]{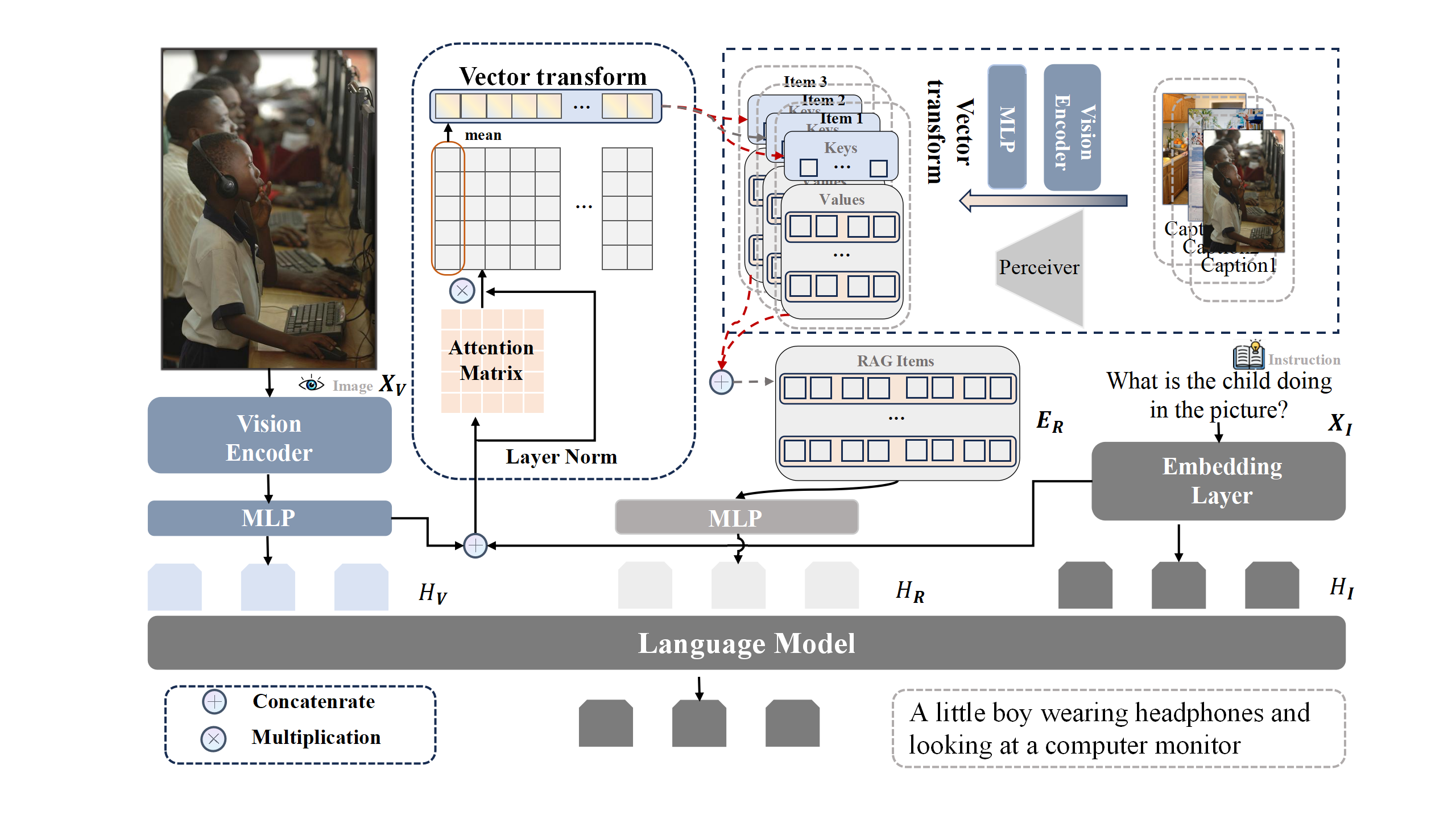} 
  \caption{Architectural overview of TinyAlign. Given an input image \(X_V\) and instruction \(X_I\), a query key derived from these inputs retrieves \(k\) similar, Perceiver-compressed multimodal embeddings \(E_R = \{E_{R_j}\}_{j=1}^k\) from a pre-constructed Memory Bank. These cues \(E_R\) are processed by a trainable RAG Connector (\(\theta_{RC}^*\)) into an auxiliary representation \(H_R\). Concurrently, \(X_V\) is processed by a Vision Transformer (ViT) and a primary Connector (\(\theta_C^*\)) into visual features \(H_V\). The instruction \(X_I\) is embedded as \(H_I\). Finally, a frozen LLM receives the composite input \(H_{\text{in}}' = [H_V, H_R, H_I]\) for prediction. This architecture enhances lightweight VLMs by supplying efficiently processed, relevant contextual information, thereby alleviating the LLM's processing burden.}
  \label{fig:tinyalign_architecture}
\end{figure}

\subsection{Theoretical Analysis: Enhancing Effective Mutual Information via RAG}
\label{subsec:theoretical_analysis_rag}
Lightweight, frozen LLMs (\(\theta_{\text{LLM,small}}\)) often exhibit high irreducible error \(\bar{\epsilon}_{\theta_{\text{LLM}}}\). TinyAlign (Fig.~\ref{fig:tinyalign_architecture}), a Retrieval-Augmented Generation (RAG)-enhanced connector architecture, mitigates this by boosting \textit{effective mutual information} (\(\Ieff\)) with strategically compressed, relevant contextual cues, thereby reducing the LLM's intrinsic processing demands.

A standard VLM maps visual input \(X_V\) (via ViT \(\theta_{\text{ViT}}\) and primary connector \(\theta_C^*\)) to \(H_V\), with instruction embeddings \(H_I\). TinyAlign augments this by: 1) retrieving \(k\) pre-compressed embeddings \(E_R\) from a memory bank \(\mathcal{M}\) (Sec.~\ref{subsec:memory_design}); 2) employing a trainable RAG connector (\(\theta_{RC}^*\)) to transform \(E_R\) into supplementary representations \(H_R\); and 3) presenting a composite input \(H_{\text{in}}' = [H_V, H_R, H_I]\) to the frozen LLM. Trainable parameters are \(\Theta_C^* = \{\theta_C^*, \theta_{RC}^*\}\).
We posit that incorporating \(E_R\) (forming augmented context \(X' = (X_V,  E_R, X_I)\)) enhances \(\Ieff\). The change, \(\Delta \Ieff\), is:
\begin{align}
\Delta \Ieff &= [I(X'; L) - \bar{\epsilon}_{\theta_{\text{LLM}}}(X')] - [I(X_V, X_I; L) - \bar{\epsilon}_{\theta_{\text{LLM}}}(X_V, X_I)] \nonumber \\
&= \underbrace{I(E_R; L | X_V, X_I)}_{\Delta I_{\text{true}}} + \underbrace{(\bar{\epsilon}_{\theta_{\text{LLM}}}(X_V, X_I) - \bar{\epsilon}_{\theta_{\text{LLM}}}(X'))}_{\Delta \bar{\epsilon}_{\theta_{\text{LLM}}}} \label{eq:delta_ieff_rag_revised_method_compressed_v2}
\end{align}
\(\Delta I_{\text{true}} > 0\) as \(E_R\) (from pertinent captions) provides novel information about target \(L\). \(\Delta \bar{\epsilon}_{\theta_{\text{LLM}}} > 0\) signifies reduced LLM irreducible error. The RAG connector \(\theta_{RC}^*\) transforms \(E_R\) into 'LLM-assimilable contextual hints' from pre-compressed image-text pairs (\(\theta_P\)-processed), making \(H_{\text{in}}'\) more 'input-friendly'. This enables the fixed-capacity LLM to better approximate \(\Ptrue(L|X')\), reducing misinterpretations (\(\bar{\epsilon}_{\theta_{\text{LLM}}}(X') < \bar{\epsilon}_{\theta_{\text{LLM}}}(X_V, X_I)\)) and alleviating cognitive load, corroborated by faster convergence (Fig.~\ref{fig:combined_loss_and_umap}(a)),For more theoretical details on introducing RAG to enhance effective mutual information,  please refer to Appendix~\ref{sec:appendix_theory}.

This \(\Delta \bar{\epsilon}_{\theta_{\text{LLM}}}\) reduction is crucial for lightweight VLMs with high baseline error. TinyAlign's architecture (LLM-independent, Perceiver-based pre-compression and efficient RAG connector) maximizes this reduction, acting as a 'cognitive scaffold' and lowering the reasoning threshold. Consequently, when \(\Delta \Ieff > 0\) (driven by \(\Delta I_{\text{true}} > 0\) and substantial \(\Delta \bar{\epsilon}_{\theta_{\text{LLM}}} > 0\)), the minimum achievable CE loss is reduced: \(\min_{\theta_C^*} \LCE(\text{RAG-enhanced}) < \min_{\theta_C^*} \LCE(\text{standard})\). RAG-enhanced connectors elevate VLM performance by mitigating the burden on frozen (especially lightweight) LLMs using pre-compressed cues, increasing leverageable information despite fixed LLM capacity.

\subsection{Memory Bank Design for Lightweight Efficiency}
\label{subsec:memory_design}
TinyAlign's memory bank \(\mathcal{M}\) supports lightweight VLMs by storing extensive multimodal information as compact key-value \((K_m, V_m)\) pairs from a large-scale pre-training dataset, prioritizing minimal storage and low retrieval latency, and we sample 100K image-text pairs from  pre-training dataset declared in Sec.~\ref{subsec:experimental_setup}.
\paragraph{Key Generation.}
Each key \(K_m \in \mathbb{R}^{d_k}\) constitutes a compact, low-dimensional embedding derived from a source image-text pair \((X_{V_m}, X_{I_m})\). Its generation employs an attention-based aggregation mechanism. Specifically, multi-modal input features first pass through a self-attention layer to yield an attention matrix, capturing contextual dependencies within the inputs. These original features are then element-wise modulated by this attention matrix. Subsequently, the resultant attention-weighted features are subjected to a mean pooling operation across the token dimension('Vector transform' in Fig.~\ref{fig:tinyalign_architecture}), producing the distilled key vector \(K_m\). This process is designed to distill salient cross-modal information into a dense representation, thereby facilitating efficient similarity search and establishing \(K_m\) as a unified query anchor.

\paragraph{Value Generation.} Each value \(V_m \in \mathbb{R}^{d_v}\) is the corresponding compressed latent multimodal embedding \(E_{R_m}\). An LLM-independent Perceiver\cite{jaegle2022perceiveriogeneralarchitecture} model (\(\theta_P\), Table~\ref{tab:perceiver_config})  pre-processes original multimodal instances \(\{ (X_{V_j}, \text{Cap}_j) \}\) into these values. This pre-compression optimizes for inference speed and reduced memory transfer.

This dual-compression strategy (compact keys \(K_m\) for rapid search, highly compressed values \(V_m\) for minimal overhead) is central to TinyAlign's efficiency, providing potent contextual cues with parsimonious resource use.

\subsection{Integrated Pre-training and Instruction Tuning Stages}
\label{subsec:training_stages}
TinyAlign employs a two-stage training strategy, with both stages using the same objective from Eq.~\eqref{eq:min_loss_ieff_vlm}: initial connector pre-training with a frozen lightweight LLM and ViT, followed by joint LLM and connector fine-tuning with a frozen ViT for task adaptation.

\textbf{Pre-training Stage.} Connectors \(\theta_C^*\) and \(\theta_{RC}^*\) are trained while the vision encoder (\(\theta_{\text{ViT}}\)) and LLM (\(\theta_{\text{LLM,small}}\)) remain frozen. This compels connectors to optimally format retrieved \(E_R\) (forming augmented input \(H_{\text{in}}' = [H_V, H_R, H_I]\)) for the LLM's extant capabilities, addressing its higher irreducible error and reducing cognitive load. This phase validates the connector's ability to map the semantic spaces of different pre-trained models to the LLM's semantic space.

\textbf{Instruction Tuning Stage.} This stage adapts the VLM to downstream tasks. The ViT (\(\theta_{\text{ViT}}\)) remains frozen. Connectors (\(\theta_C^*, \theta_{RC}^*\)) and, critically, the lightweight LLM (\(\theta_{\text{LLM,small}}\)) are fine-tuned. With connectors supplying simplified, relevant input, LLM fine-tuning enables specialized reasoning by maximally leveraging these augmented inputs from an active memory bank \(\mathcal{M}\). This integrated approach optimizes the LLM's utilization of the enhanced inputs.

\section{Experiments}
\label{sec:experiments}
\begin{figure}[htbp]
  \centering
  \begin{minipage}[t]{0.48\textwidth}
    \centering
    \includegraphics[width=\linewidth]{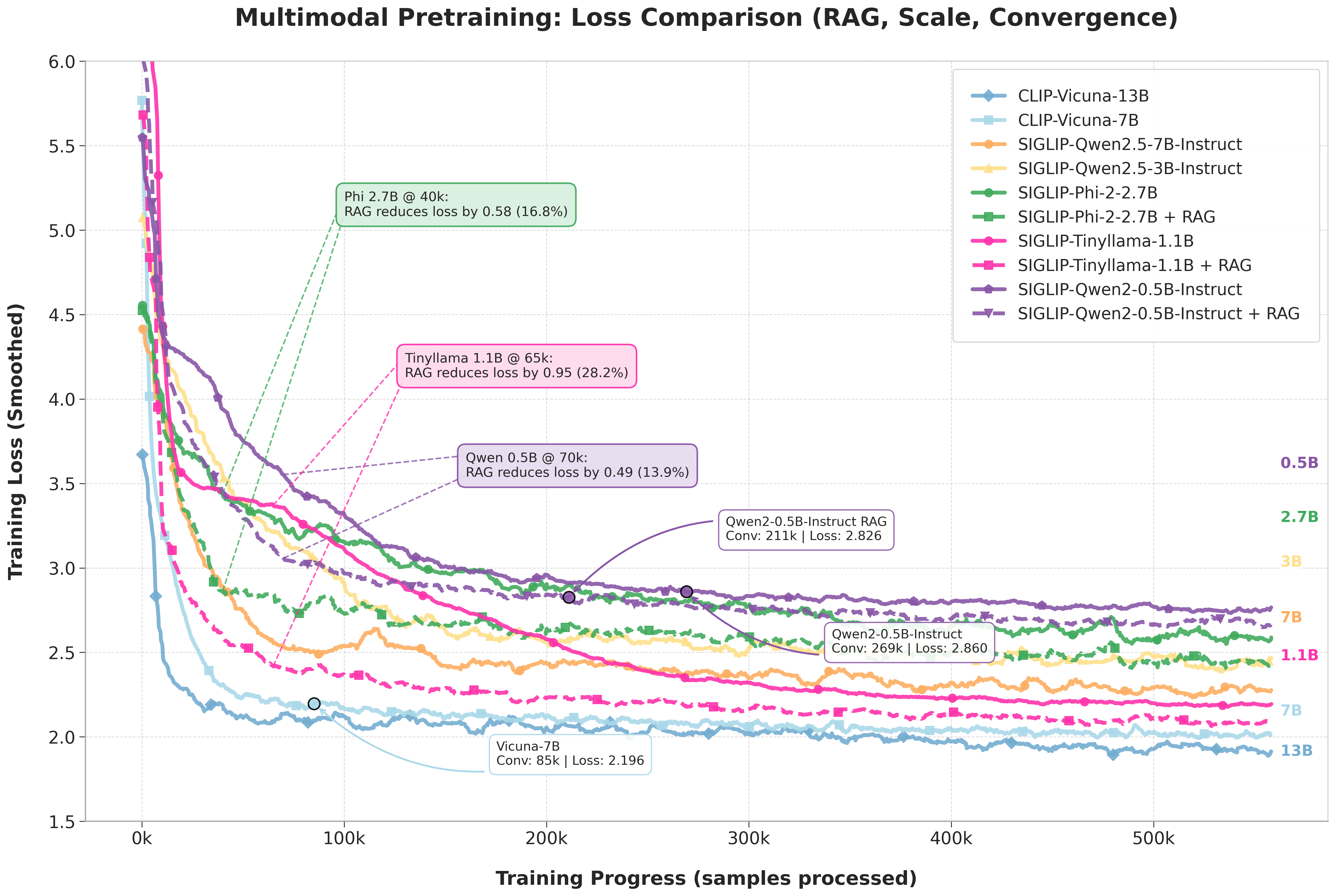} 
    \vspace{0.5ex} 
    (a) Pre-training Loss Dynamics
  \end{minipage}\hfill
  \begin{minipage}[t]{0.48\textwidth}
    \centering
    \includegraphics[width=\linewidth]{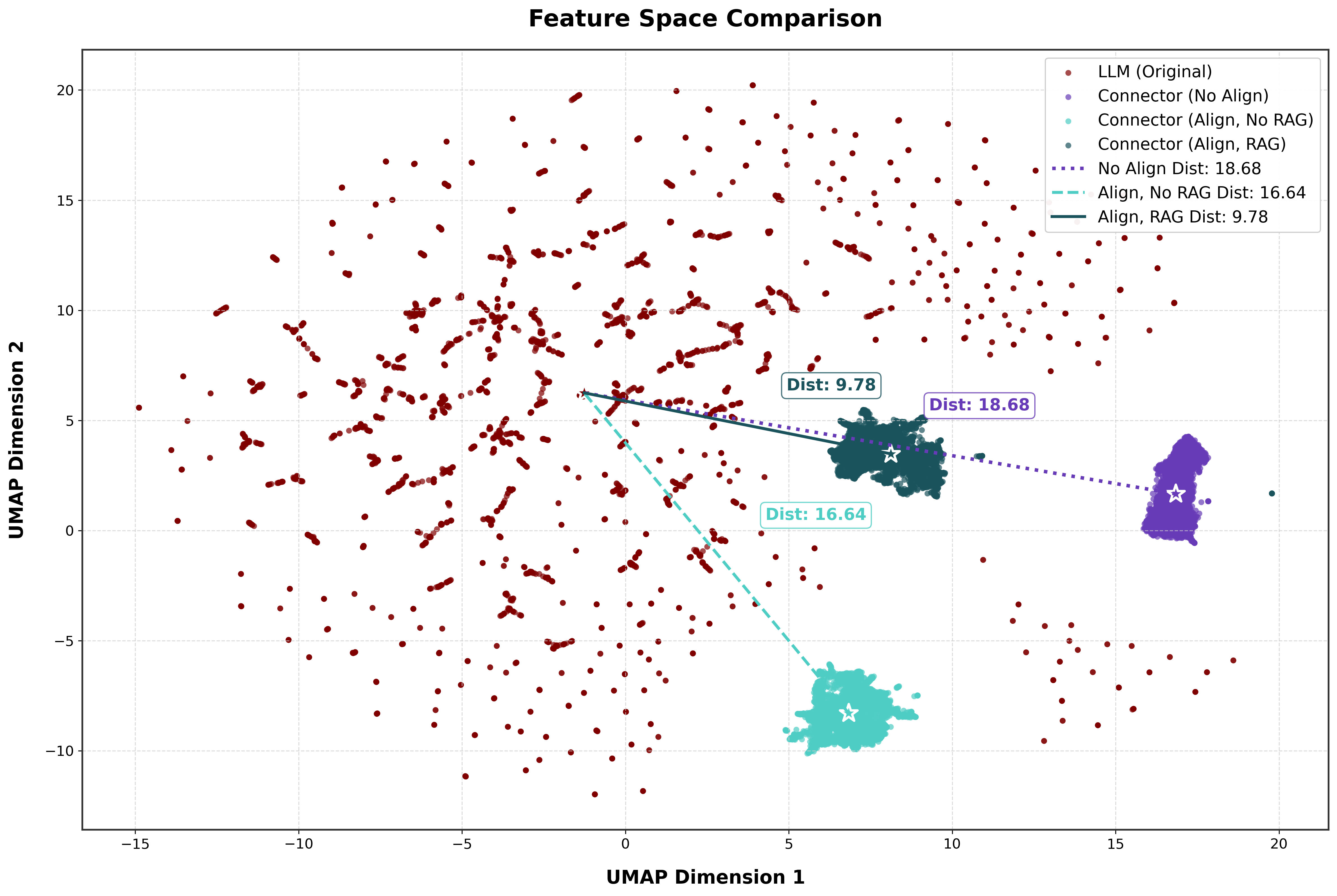} 
    \vspace{0.5ex} 
    (b) UMAP of LLM Input Space (Phi-2)
  \end{minipage}
  \caption{\textbf{(a)} Comparison of multimodal pre-training loss on the LLaVA dataset. TinyAlign-enhanced models (dashed lines, "+ RAG") exhibit accelerated convergence and lower final training loss compared to baselines (solid lines) across various model scales (e.g., 16.8\% loss reduction for Phi-2.7B, 28.2\% for TinyLLaMA-1.1B, 13.9\% for Qwen2-0.5B at specified sample points). All models use SigLIP vision encoders, with only connector parameters trained. \textbf{(b)} UMAP visualization of the LLM (for example, Phi-2) input space for (i) baseline(dashed lines), (ii) TinyAlign-enhanced (solid lines), and (iii) a non-aligned model. TinyAlign promotes superior semantic clustering and alignment.}
  \label{fig:combined_loss_and_umap}
\end{figure}

We conduct a comprehensive suite of experiments to rigorously validate the efficacy of TinyAlign. Our evaluation is designed to: (1) assess its impact on pre-training dynamics, specifically convergence speed and loss reduction (Fig.~\ref{fig:combined_loss_and_umap}(a)); (2) analyze the quality of the learned representations through UMAP visualizations (Fig.~\ref{fig:combined_loss_and_umap}(b)); (3) quantify performance improvements on diverse downstream multimodal tasks post-instruction tuning, detailed in Table~\ref{tab:performance_summary}; and (4) measure gains in data efficiency during instruction tuning (Sec.~\ref{subsec:data_efficiency_condensed}, Fig.~\ref{fig:data_efficiency_main}), A thorough analysis of data efficiency across various benchmarks can be found in Appendix~\ref{sec:appendix_detailed_data_efficiency} (Fig.~\ref{fig:bech_appendix}). Additionally, our FLOPs analysis, detailed in Appendix~\ref{sec:appendix_flops_analysis} (Table~\ref{tab:flops_analysis_table}). The extensive ablation studies validating our architectural decisions are presented in Appendix~\ref{sec:appendix_detailed_Ablation_Studies} (Tables~\ref{tab:ablation_kb_size_appendix}, \ref{tab:ablation_vit_alignment_appendix}, and \ref{tab:ablation_top_k_phi_appendix}). These experiments empirically substantiate our theoretical derivations concerning the enhancement of Effective Mutual Information (\(\Ieff\)). 

\subsection{Experimental Setup}
\label{subsec:experimental_setup}
Our framework builds upon lightweight VLMs, drawing inspiration from TinyLLaVA \cite{zhou2024tinyllavaframeworksmallscalelarge}. We primarily utilize the LLaVA dataset family \cite{liu2023visualinstructiontuning} for pre-training analysis and Instrcution tuning, Table~\ref{tab:hyperparameters} provides details on the computational key hyperparameters used for the pre-training and Instruction tuning phases of our experiments. Detailed hyperparameter settings are available in Appendix~\ref{sec:appendix_hyperparameters}.
\paragraph{Datasets.}
\textbf{Pre-training Data:} For analyzing pre-training dynamics (Sec.~\ref{subsec:pretrain_performance}) and constructing the memory bank (details in Appendix~\ref{sec:appendix_detailed_Ablation_Studies}), we employ the LLaVA pre-training set. This set comprises 558K image-text pairs from a LAION-CC-SBU subset \cite{liu2023visualinstructiontuning}, with corresponding annotations.
\textbf{Instruction Tuning Data:} For evaluating instruction-following capabilities, we use the LLaVA v1.5 SFT dataset, containing approximately 665K samples. Images are sourced from diverse datasets including COCO \cite{lin2015microsoftcococommonobjects}, GQA \cite{hudson2019gqanewdatasetrealworld}, TextVQA \cite{singh2019vqamodelsread}, OCR-VQA \cite{mishraICDAR19}, and VisualGenome \cite{krishna2016visualgenomeconnectinglanguage}.
\paragraph{Models and Baselines.}
We validate TinyAlign across several lightweight Large Language Models (LLMs) and vision encoders: Vicuna-7B/13B \cite{vicuna2023}, Phi-2 (2.7B) \cite{gunasekar2023textbooksneed}, Gemma (2B) \cite{gemmateam2024gemma2improvingopen}, TinyLLaMA (1.1B) \cite{zhang2024tinyllamaopensourcesmalllanguage}, and Qwen2-0.5B \cite{bai2025qwen25vltechnicalreport}. For vision encoders, we use SigLIP \cite{zhai2023sigmoidlosslanguageimage} and CLIP \cite{radford2021learningtransferablevisualmodels}. Consistent with findings in TinyLLaVA \cite{zhou2024tinyllavaframeworksmallscalelarge} that favor SigLIP for lightweight VLMs, we primarily adopt SigLIP. The Connectors utilize a two-layer multilayer perceptron with GELU activation functions to transform the input features.

\subsection{Pre-training Performance Analysis}
\label{subsec:pretrain_performance}
Our empirical results demonstrate TinyAlign's substantial efficacy in enhancing pre-training for lightweight Vision-Language Models (VLMs). Analysis of training loss trajectories and learned multimodal representations reveals two key benefits.

\textbf{Accelerated Convergence and Reduced Training Loss.}
TinyAlign markedly accelerates convergence and reduces final training loss on the LLaVA pre-training dataset. As shown in Fig.~\ref{fig:combined_loss_and_umap}(a), models incorporating TinyAlign, such as Phi-2 (2.7B), TinyLLaMA (1.1B), and Qwen2-0.5B, achieve significant loss reductions of 16.8\%, 28.2\%, and 13.9\%, respectively, compared to baselines. This improved training efficiency and superior model fit empirically validate our hypothesis (Eq.~\eqref{eq:min_loss_constrained_vlm}) that TinyAlign enhances Effective Mutual Information (\(\Ieff\)), thereby lowering the achievable cross-entropy loss (\(\mathcal{L}_{\text{CE}}\)). These benefits are consistent across models ranging from 0.5B to 2.7B parameters, underscoring TinyAlign's broad applicability.

\textbf{Improved Embedding Space Alignment and Semantic Coherence.}
Qualitatively, TinyAlign cultivates superior multimodal representations.
UMAP projections (Fig.~\ref{fig:combined_loss_and_umap}(b)) reveal that TinyAlign (ii) produces significantly tighter, more separable, and semantically coherent image-text embedding clusters compared to the diffuse embeddings from baselines (i) and the disarray in non-aligned models (iii).
This improved structural alignment, indicative of a reduced average embedding error ($\bar{\epsilon}_{\theta_{\text{LLM}}}$), is attributed to contextual cues $H_R$ from retrieved information guiding the LLM towards more accurate inter-modal associations.
Such well-structured representations are crucial for downstream task performance (see Appendix ~\ref{sec:appendix_umap} for UMAP details).

In summary, TinyAlign demonstrably improves pre-training efficiency and yields higher-quality multimodal representations. These advancements establish a more robust foundation for downstream instruction tuning and enhanced performance across diverse VLM applications.

\subsection{Instruction Tuning Performance Analysis}
\label{subsec:finetune_performance}
 Post-instruction tuning, TinyAlign variants consistently outperform their respective baselines across a majority of the evaluated benchmarks, as detailed in Table~\ref{tab:performance_summary}. For instance, TinyLLaMA-1.1B equipped with TinyAlign achieves a 74.49\% accuracy on VQAv2, marking a 3.51\% improvement over its baseline. Similarly, Phi-2-2.7B with TinyAlign attains 78.32\% on VQAv2, a 2.93\% gain. These improvements generally underscore enhanced generalization capabilities stemming from superior pre-training alignment. However, on the highly complex MM-Vet benchmark, the TinyLLaMA-1.1B variant with TinyAlign exhibits a marginal performance decrease (2.4\%). This suggests that for certain exceptionally challenging tasks, particularly with smaller model capacities, the benefits of alignment might be nuanced or necessitate further task-specific tuning.

\begin{table}[htbp]
\centering
\caption{Performance comparison on various multimodal benchmarks after instruction tuning. \uparr indicates performance improvement with TinyAlign, while \downarr indicates a decrease.}
\label{tab:performance_summary} 
\resizebox{\textwidth}{!}{%
\begin{tabular}{@{}l cc cc cc cc@{}}
\toprule
\multirow{2}{*}{Benchmark} & \multicolumn{2}{c}{TinyLLaMA-1.1B} & \multicolumn{2}{c}{Phi-2-2.7B} & \multicolumn{2}{c}{Qwen2-0.5B} \\
\cmidrule(lr){2-3} \cmidrule(lr){4-5} \cmidrule(lr){6-7} \cmidrule(lr){8-9}
& Baseline & +TinyAlign & Baseline & +TinyAlign & Baseline & +TinyAlign \\
\midrule
GQA \cite{hudson2019gqanewdatasetrealworld}      & 52.37 & 56.69 \uparr  & 58.44 & 60.73 \uparr & 56.30 & 57.58 \uparr\\
MMMU \cite{yue2024mmmumassivemultidisciplinemultimodal}     & 29.4  & 30.2 \uparr   & 36.2 & 37.7 \uparr & 31.0 & 31.4 \uparr\\
MM-Vet \cite{yu2024mmvetevaluatinglargemultimodal}   & 26.0  & 23.6 \downarr & 31.5 & 34.1 \uparr  & 20.9 & 23.6 \uparr \\
POPE \cite{li2023evaluatingobjecthallucinationlarge}    & 84.3  & 86.0 \uparr   & 86.6 & 88.0 \uparr & 86.4 & 87.2 \uparr \\
SQA-I \cite{lu2022learnexplainmultimodalreasoning}    & 56.52 & 59.84 \uparr  & 67.28 & 68.12 \uparr & 59.54 & 60.24 \uparr \\
TextVQA \cite{singh2019vqamodelsread}  & 46.33 & 46.41 \uparr  & 50.31 & 55.48 \uparr & 46.08 & 46.62 \uparr  \\
VQAV2 \cite{agrawal2016vqavisualquestionanswering}    & 70.98 & 74.49 \uparr  & 75.39 & 78.32 \uparr & 72.95 & 74.21 \uparr  \\
MME \cite{fu2024mmecomprehensiveevaluationbenchmark}    & 1105.10 & 1200.67 \uparr & 1364.2 & 1412.1 \uparr & 1170.54 & 1208.87 \uparr  \\
\bottomrule
\end{tabular}%
}
\end{table}
\subsection{Data Efficiency Analysis}
\label{subsec:data_efficiency_condensed}
\begin{wrapfigure}{r}{0.5\textwidth}
  \centering
  \vspace{-\intextsep} 
  \includegraphics[width=1.0\linewidth]{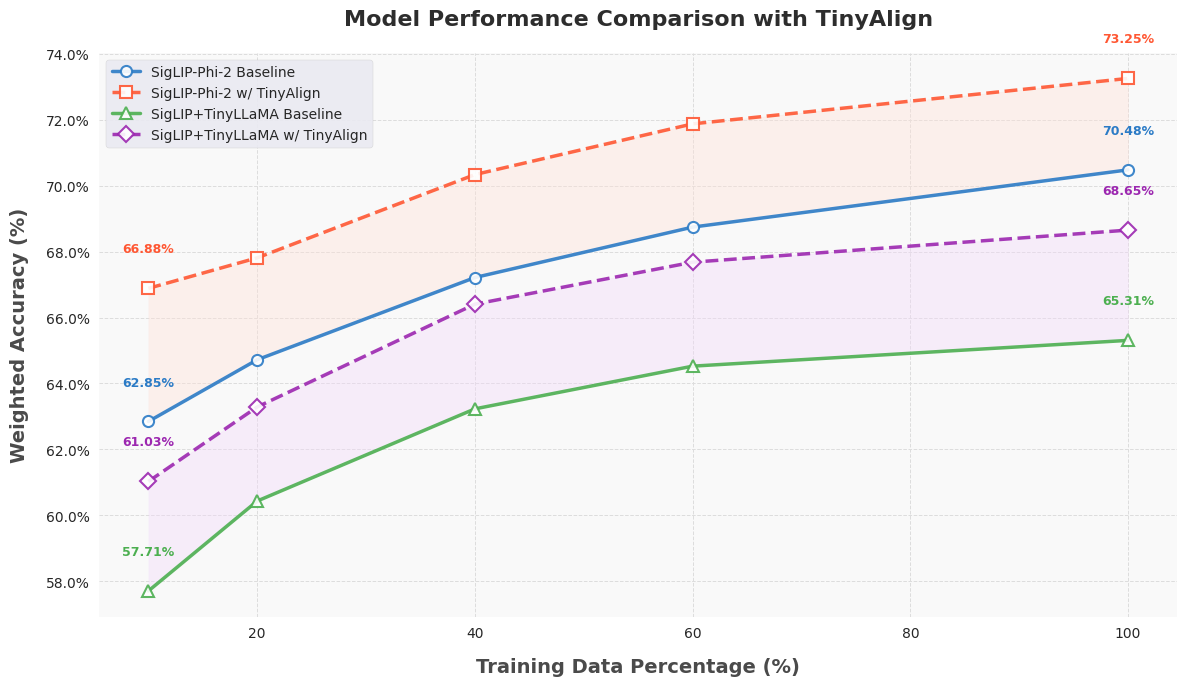} 
  \caption{Model weighted accuracy, computed per Eq.~\eqref{eq:acc}, is evaluated against the percentage of instruction tuning data utilized. TinyAlign models consistently demonstrate superior performance, achieving higher accuracy with substantially less data relative to baseline models.}
  \label{fig:data_efficiency_main} 
  \vspace{-4mm} 
\end{wrapfigure}

TinyAlign demonstrates substantial improvements in data efficiency during instruction tuning, as depicted in Fig.~\ref{fig:data_efficiency_main}. Our analysis reveals that TinyAlign-enhanced models (Phi-2 and TinyLLaMA) can achieve a weighted average accuracy comparable to baselines trained on 100\% of the instruction tuning data while utilizing only 40\% of the data. This efficiency gain is observed across various capabilities, as the Fig.~\ref{fig:bech_appendix} shows: TinyAlign significantly enhances visual reasoning (e.g., GQA, VQAv2, SQA-I), improves scene text understanding (e.g., TextVQA), and curtails model hallucinations (e.g., POPE), frequently matching or exceeding baseline performance with merely 20-60\% of the data. While challenges persist for highly complex tasks such as MMMU, where gains are less uniform, the overall results compellingly underscore TinyAlign's capacity to facilitate the development of highly capable VLMs under data-constrained scenarios. This reduces reliance on extensive labeled datasets and can accelerate development cycles. A detailed benchmark-specific data efficiency analysis is provided in Appendix~\ref{sec:appendix_detailed_data_efficiency}.

\subsection{Ablation Study}
\label{subsec:ablation_studies_main} 
To systematically assess  our model architecture, we conduct a series of targeted ablation studies focusing on three key design choices: (1) the size and composition of the RAG knowledge base (Memory Bank) (Table~\ref{tab:ablation_kb_size_appendix}), (2) the design methodology for the knowledge base keys (Table~\ref{tab:ablation_vit_alignment_appendix}), and (3) the number of documents (Top-K) retrieved during inference (Table~\ref{tab:ablation_top_k_phi_appendix}).

The findings from these ablations, detailed in Appendix~\ref{sec:appendix_detailed_Ablation_Studies}, inform our final model configuration. We adopt a 100k-entry knowledge base, ensure alignment between the vision encoders used for VLM training and RAG key generation, and retrieve the top-5 documents to augment generation. These choices strike an optimal balance of performance, efficiency, and contextual enrichment.

\section{Discussion}
\label{sec:conclusion}

\textbf{Conclusion.} Lightweight Vision-Language Models  are essential for resource-constrained applications, yet their performance is often limited by alignment bottlenecks caused by the constrained capacity of smaller language models. Through a mutual information perspective, we demonstrate that this limitation reduces the Effective Mutual Information between multimodal inputs and outputs, compromising alignment quality. To address this, we propose TinyAlign, a framework inspired by Retrieval-Augmented Generation, which retrieves relevant context from a memory bank to enrich multimodal inputs. Empirical evaluations show that TinyAlign significantly improves task performance, reduces training loss, and accelerates convergence. Notably, it achieves baseline-level performance with only 40\% of the fine-tuning data, offering a highly data-efficient solution for lightweight VLMs. This work provides both a practical framework and a theoretical foundation to address alignment challenges in constrained multimodal systems.

\textbf{Limitation and Future Work.} Though TinyAlign is effective, it has several limitations. Its reliance on a well-designed memory bank may require bespoke construction for different LLMs, and compatibility between the Vision Transformer or querying and memory key generation is essential. Additionally, for challenging tasks or smaller models, further task-specific tuning may be needed. Future work could explore adaptive memory construction and more flexible ViT compatibility to enhance alignment across diverse settings.

\newpage
\bibliographystyle{plainnat}
\bibliography{references}  

\begin{thebibliography}{47}
\providecommand{\natexlab}[1]{#1}
\providecommand{\url}[1]{\texttt{#1}}
\expandafter\ifx\csname urlstyle\endcsname\relax
  \providecommand{\doi}[1]{doi: #1}\else
  \providecommand{\doi}{doi: \begingroup \urlstyle{rm}\Url}\fi

\bibitem[Agrawal et~al.(2016)Agrawal, Lu, Antol, Mitchell, Zitnick, Batra, and Parikh]{agrawal2016vqavisualquestionanswering}
Aishwarya Agrawal, Jiasen Lu, Stanislaw Antol, Margaret Mitchell, C.~Lawrence Zitnick, Dhruv Batra, and Devi Parikh.
\newblock Vqa: Visual question answering, 2016.
\newblock URL \url{https://arxiv.org/abs/1505.00468}.

\bibitem[Bai et~al.(2025)Bai, Chen, Liu, Wang, Ge, Song, Dang, Wang, Wang, Tang, Zhong, Zhu, Yang, Li, Wan, Wang, Ding, Fu, Xu, Ye, Zhang, Xie, Cheng, Zhang, Yang, Xu, and Lin]{bai2025qwen25vltechnicalreport}
Shuai Bai, Keqin Chen, Xuejing Liu, Jialin Wang, Wenbin Ge, Sibo Song, Kai Dang, Peng Wang, Shijie Wang, Jun Tang, Humen Zhong, Yuanzhi Zhu, Mingkun Yang, Zhaohai Li, Jianqiang Wan, Pengfei Wang, Wei Ding, Zheren Fu, Yiheng Xu, Jiabo Ye, Xi~Zhang, Tianbao Xie, Zesen Cheng, Hang Zhang, Zhibo Yang, Haiyang Xu, and Junyang Lin.
\newblock Qwen2.5-vl technical report, 2025.
\newblock URL \url{https://arxiv.org/abs/2502.13923}.

\bibitem[Caffagni et~al.(2024)Caffagni, Cocchi, Moratelli, Sarto, Cornia, Baraldi, and Cucchiara]{caffagni2024wikillavahierarchicalretrievalaugmentedgeneration}
Davide Caffagni, Federico Cocchi, Nicholas Moratelli, Sara Sarto, Marcella Cornia, Lorenzo Baraldi, and Rita Cucchiara.
\newblock Wiki-llava: Hierarchical retrieval-augmented generation for multimodal llms, 2024.
\newblock URL \url{https://arxiv.org/abs/2404.15406}.

\bibitem[Chen et~al.(2023{\natexlab{a}})Chen, Zhu, Shen, Li, Liu, Zhang, Krishnamoorthi, Chandra, Xiong, and Elhoseiny]{chen2023minigptv2largelanguagemodel}
Jun Chen, Deyao Zhu, Xiaoqian Shen, Xiang Li, Zechun Liu, Pengchuan Zhang, Raghuraman Krishnamoorthi, Vikas Chandra, Yunyang Xiong, and Mohamed Elhoseiny.
\newblock Minigpt-v2: large language model as a unified interface for vision-language multi-task learning, 2023{\natexlab{a}}.
\newblock URL \url{https://arxiv.org/abs/2310.09478}.

\bibitem[Chen et~al.(2022)Chen, Hu, Chen, Verga, and Cohen]{chen2022muragmultimodalretrievalaugmentedgenerator}
Wenhu Chen, Hexiang Hu, Xi~Chen, Pat Verga, and William~W. Cohen.
\newblock Murag: Multimodal retrieval-augmented generator for open question answering over images and text, 2022.
\newblock URL \url{https://arxiv.org/abs/2210.02928}.

\bibitem[Chen et~al.(2023{\natexlab{b}})Chen, Djolonga, Padlewski, Mustafa, Changpinyo, Wu, Ruiz, Goodman, Wang, Tay, Shakeri, Dehghani, Salz, Lucic, Tschannen, Nagrani, Hu, Joshi, Pang, Montgomery, Pietrzyk, Ritter, Piergiovanni, Minderer, Pavetic, Waters, Li, Alabdulmohsin, Beyer, Amelot, Lee, Steiner, Li, Keysers, Arnab, Xu, Rong, Kolesnikov, Seyedhosseini, Angelova, Zhai, Houlsby, and Soricut]{chen2023palixscalingmultilingualvision}
Xi~Chen, Josip Djolonga, Piotr Padlewski, Basil Mustafa, Soravit Changpinyo, Jialin Wu, Carlos~Riquelme Ruiz, Sebastian Goodman, Xiao Wang, Yi~Tay, Siamak Shakeri, Mostafa Dehghani, Daniel Salz, Mario Lucic, Michael Tschannen, Arsha Nagrani, Hexiang Hu, Mandar Joshi, Bo~Pang, Ceslee Montgomery, Paulina Pietrzyk, Marvin Ritter, AJ~Piergiovanni, Matthias Minderer, Filip Pavetic, Austin Waters, Gang Li, Ibrahim Alabdulmohsin, Lucas Beyer, Julien Amelot, Kenton Lee, Andreas~Peter Steiner, Yang Li, Daniel Keysers, Anurag Arnab, Yuanzhong Xu, Keran Rong, Alexander Kolesnikov, Mojtaba Seyedhosseini, Anelia Angelova, Xiaohua Zhai, Neil Houlsby, and Radu Soricut.
\newblock Pali-x: On scaling up a multilingual vision and language model, 2023{\natexlab{b}}.
\newblock URL \url{https://arxiv.org/abs/2305.18565}.

\bibitem[Chen et~al.(2024)Chen, Wu, Wang, Su, Chen, Xing, Zhong, Zhang, Zhu, Lu, Li, Luo, Lu, Qiao, and Dai]{chen2024internvlscalingvisionfoundation}
Zhe Chen, Jiannan Wu, Wenhai Wang, Weijie Su, Guo Chen, Sen Xing, Muyan Zhong, Qinglong Zhang, Xizhou Zhu, Lewei Lu, Bin Li, Ping Luo, Tong Lu, Yu~Qiao, and Jifeng Dai.
\newblock Internvl: Scaling up vision foundation models and aligning for generic visual-linguistic tasks, 2024.
\newblock URL \url{https://arxiv.org/abs/2312.14238}.

\bibitem[Chiang et~al.(2023)Chiang, Li, Lin, Sheng, Wu, Zhang, Zheng, Zhuang, Zhuang, Gonzalez, Stoica, and Xing]{vicuna2023}
Wei-Lin Chiang, Zhuohan Li, Zi~Lin, Ying Sheng, Zhanghao Wu, Hao Zhang, Lianmin Zheng, Siyuan Zhuang, Yonghao Zhuang, Joseph~E. Gonzalez, Ion Stoica, and Eric~P. Xing.
\newblock Vicuna: An open-source chatbot impressing gpt-4 with 90\%* chatgpt quality, March 2023.
\newblock URL \url{https://lmsys.org/blog/2023-03-30-vicuna/}.

\bibitem[Chu et~al.(2023)Chu, Qiao, Lin, Xu, Yang, Hu, Wei, Zhang, Zhang, Wei, and Shen]{chu2023mobilevlmfaststrong}
Xiangxiang Chu, Limeng Qiao, Xinyang Lin, Shuang Xu, Yang Yang, Yiming Hu, Fei Wei, Xinyu Zhang, Bo~Zhang, Xiaolin Wei, and Chunhua Shen.
\newblock Mobilevlm : A fast, strong and open vision language assistant for mobile devices, 2023.
\newblock URL \url{https://arxiv.org/abs/2312.16886}.

\bibitem[Chu et~al.(2024)Chu, Qiao, Zhang, Xu, Wei, Yang, Sun, Hu, Lin, Zhang, and Shen]{chu2024mobilevlmv2fasterstronger}
Xiangxiang Chu, Limeng Qiao, Xinyu Zhang, Shuang Xu, Fei Wei, Yang Yang, Xiaofei Sun, Yiming Hu, Xinyang Lin, Bo~Zhang, and Chunhua Shen.
\newblock Mobilevlm v2: Faster and stronger baseline for vision language model, 2024.
\newblock URL \url{https://arxiv.org/abs/2402.03766}.

\bibitem[Fu et~al.(2024)Fu, Chen, Shen, Qin, Zhang, Lin, Yang, Zheng, Li, Sun, Wu, and Ji]{fu2024mmecomprehensiveevaluationbenchmark}
Chaoyou Fu, Peixian Chen, Yunhang Shen, Yulei Qin, Mengdan Zhang, Xu~Lin, Jinrui Yang, Xiawu Zheng, Ke~Li, Xing Sun, Yunsheng Wu, and Rongrong Ji.
\newblock Mme: A comprehensive evaluation benchmark for multimodal large language models, 2024.
\newblock URL \url{https://arxiv.org/abs/2306.13394}.

\bibitem[{Google}(2024)]{gemini2.5pro2024}
{Google}.
\newblock Gemini 2.5 pro.
\newblock \url{https://deepmind.google/technologies/gemini/}, 2024.
\newblock Google DeepMind.

\bibitem[Gunasekar et~al.(2023)Gunasekar, Zhang, Aneja, Mendes, Giorno, Gopi, Javaheripi, Kauffmann, de~Rosa, Saarikivi, Salim, Shah, Behl, Wang, Bubeck, Eldan, Kalai, Lee, and Li]{gunasekar2023textbooksneed}
Suriya Gunasekar, Yi~Zhang, Jyoti Aneja, Caio César~Teodoro Mendes, Allie~Del Giorno, Sivakanth Gopi, Mojan Javaheripi, Piero Kauffmann, Gustavo de~Rosa, Olli Saarikivi, Adil Salim, Shital Shah, Harkirat~Singh Behl, Xin Wang, Sébastien Bubeck, Ronen Eldan, Adam~Tauman Kalai, Yin~Tat Lee, and Yuanzhi Li.
\newblock Textbooks are all you need, 2023.
\newblock URL \url{https://arxiv.org/abs/2306.11644}.

\bibitem[Guu et~al.(2020)Guu, Lee, Tung, Pasupat, and Chang]{guu2020realmretrievalaugmentedlanguagemodel}
Kelvin Guu, Kenton Lee, Zora Tung, Panupong Pasupat, and Ming-Wei Chang.
\newblock Realm: Retrieval-augmented language model pre-training, 2020.
\newblock URL \url{https://arxiv.org/abs/2002.08909}.

\bibitem[Hu et~al.(2023)Hu, Iscen, Sun, Wang, Chang, Sun, Schmid, Ross, and Fathi]{hu2023revealretrievalaugmentedvisuallanguagepretraining}
Ziniu Hu, Ahmet Iscen, Chen Sun, Zirui Wang, Kai-Wei Chang, Yizhou Sun, Cordelia Schmid, David~A. Ross, and Alireza Fathi.
\newblock Reveal: Retrieval-augmented visual-language pre-training with multi-source multimodal knowledge memory, 2023.
\newblock URL \url{https://arxiv.org/abs/2212.05221}.

\bibitem[Hudson and Manning(2019)]{hudson2019gqanewdatasetrealworld}
Drew~A. Hudson and Christopher~D. Manning.
\newblock Gqa: A new dataset for real-world visual reasoning and compositional question answering, 2019.
\newblock URL \url{https://arxiv.org/abs/1902.09506}.

\bibitem[Jaegle et~al.(2022)Jaegle, Borgeaud, Alayrac, Doersch, Ionescu, Ding, Koppula, Zoran, Brock, Shelhamer, Hénaff, Botvinick, Zisserman, Vinyals, and Carreira]{jaegle2022perceiveriogeneralarchitecture}
Andrew Jaegle, Sebastian Borgeaud, Jean-Baptiste Alayrac, Carl Doersch, Catalin Ionescu, David Ding, Skanda Koppula, Daniel Zoran, Andrew Brock, Evan Shelhamer, Olivier Hénaff, Matthew~M. Botvinick, Andrew Zisserman, Oriol Vinyals, and Joāo Carreira.
\newblock Perceiver io: A general architecture for structured inputs \& outputs, 2022.
\newblock URL \url{https://arxiv.org/abs/2107.14795}.

\bibitem[Krishna et~al.(2016)Krishna, Zhu, Groth, Johnson, Hata, Kravitz, Chen, Kalantidis, Li, Shamma, Bernstein, and Li]{krishna2016visualgenomeconnectinglanguage}
Ranjay Krishna, Yuke Zhu, Oliver Groth, Justin Johnson, Kenji Hata, Joshua Kravitz, Stephanie Chen, Yannis Kalantidis, Li-Jia Li, David~A. Shamma, Michael~S. Bernstein, and Fei-Fei Li.
\newblock Visual genome: Connecting language and vision using crowdsourced dense image annotations, 2016.
\newblock URL \url{https://arxiv.org/abs/1602.07332}.

\bibitem[Lewis et~al.(2021)Lewis, Perez, Piktus, Petroni, Karpukhin, Goyal, Küttler, Lewis, tau Yih, Rocktäschel, Riedel, and Kiela]{lewis2021retrievalaugmentedgenerationknowledgeintensivenlp}
Patrick Lewis, Ethan Perez, Aleksandra Piktus, Fabio Petroni, Vladimir Karpukhin, Naman Goyal, Heinrich Küttler, Mike Lewis, Wen tau Yih, Tim Rocktäschel, Sebastian Riedel, and Douwe Kiela.
\newblock Retrieval-augmented generation for knowledge-intensive nlp tasks, 2021.
\newblock URL \url{https://arxiv.org/abs/2005.11401}.

\bibitem[Li et~al.(2023{\natexlab{a}})Li, Li, Savarese, and Hoi]{li2023blip2bootstrappinglanguageimagepretraining}
Junnan Li, Dongxu Li, Silvio Savarese, and Steven Hoi.
\newblock Blip-2: Bootstrapping language-image pre-training with frozen image encoders and large language models, 2023{\natexlab{a}}.
\newblock URL \url{https://arxiv.org/abs/2301.12597}.

\bibitem[Li et~al.(2023{\natexlab{b}})Li, Du, Zhou, Wang, Zhao, and Wen]{li2023evaluatingobjecthallucinationlarge}
Yifan Li, Yifan Du, Kun Zhou, Jinpeng Wang, Wayne~Xin Zhao, and Ji-Rong Wen.
\newblock Evaluating object hallucination in large vision-language models, 2023{\natexlab{b}}.
\newblock URL \url{https://arxiv.org/abs/2305.10355}.

\bibitem[Lin et~al.(2015)Lin, Maire, Belongie, Bourdev, Girshick, Hays, Perona, Ramanan, Zitnick, and Dollár]{lin2015microsoftcococommonobjects}
Tsung-Yi Lin, Michael Maire, Serge Belongie, Lubomir Bourdev, Ross Girshick, James Hays, Pietro Perona, Deva Ramanan, C.~Lawrence Zitnick, and Piotr Dollár.
\newblock Microsoft coco: Common objects in context, 2015.
\newblock URL \url{https://arxiv.org/abs/1405.0312}.

\bibitem[Liu et~al.(2023)Liu, Li, Wu, and Lee]{liu2023visualinstructiontuning}
Haotian Liu, Chunyuan Li, Qingyang Wu, and Yong~Jae Lee.
\newblock Visual instruction tuning, 2023.
\newblock URL \url{https://arxiv.org/abs/2304.08485}.

\bibitem[Liu et~al.(2025)Liu, Qi, He, Bisazza, Sachan, and Cotterell]{liu2025pointwisemutualinformationperformance}
Tianyu Liu, Jirui Qi, Paul He, Arianna Bisazza, Mrinmaya Sachan, and Ryan Cotterell.
\newblock Pointwise mutual information as a performance gauge for retrieval-augmented generation, 2025.
\newblock URL \url{https://arxiv.org/abs/2411.07773}.

\bibitem[Lu et~al.(2024)Lu, Liu, Zhang, Wang, Dong, Liu, Sun, Ren, Li, Yang, Sun, Deng, Xu, Xie, and Ruan]{lu2024deepseekvlrealworldvisionlanguageunderstanding}
Haoyu Lu, Wen Liu, Bo~Zhang, Bingxuan Wang, Kai Dong, Bo~Liu, Jingxiang Sun, Tongzheng Ren, Zhuoshu Li, Hao Yang, Yaofeng Sun, Chengqi Deng, Hanwei Xu, Zhenda Xie, and Chong Ruan.
\newblock Deepseek-vl: Towards real-world vision-language understanding, 2024.
\newblock URL \url{https://arxiv.org/abs/2403.05525}.

\bibitem[Lu et~al.(2022)Lu, Mishra, Xia, Qiu, Chang, Zhu, Tafjord, Clark, and Kalyan]{lu2022learnexplainmultimodalreasoning}
Pan Lu, Swaroop Mishra, Tony Xia, Liang Qiu, Kai-Wei Chang, Song-Chun Zhu, Oyvind Tafjord, Peter Clark, and Ashwin Kalyan.
\newblock Learn to explain: Multimodal reasoning via thought chains for science question answering, 2022.
\newblock URL \url{https://arxiv.org/abs/2209.09513}.

\bibitem[Marafioti et~al.(2025)Marafioti, Zohar, Farré, Noyan, Bakouch, Cuenca, Zakka, Allal, Lozhkov, Tazi, Srivastav, Lochner, Larcher, Morlon, Tunstall, von Werra, and Wolf]{marafioti2025smolvlmredefiningsmallefficient}
Andrés Marafioti, Orr Zohar, Miquel Farré, Merve Noyan, Elie Bakouch, Pedro Cuenca, Cyril Zakka, Loubna~Ben Allal, Anton Lozhkov, Nouamane Tazi, Vaibhav Srivastav, Joshua Lochner, Hugo Larcher, Mathieu Morlon, Lewis Tunstall, Leandro von Werra, and Thomas Wolf.
\newblock Smolvlm: Redefining small and efficient multimodal models, 2025.
\newblock URL \url{https://arxiv.org/abs/2504.05299}.

\bibitem[Mishra et~al.(2019)Mishra, Shekhar, Singh, and Chakraborty]{mishraICDAR19}
Anand Mishra, Shashank Shekhar, Ajeet~Kumar Singh, and Anirban Chakraborty.
\newblock Ocr-vqa: Visual question answering by reading text in images.
\newblock In \emph{ICDAR}, 2019.

\bibitem[{OpenAI}(2023)]{gpt4v2023}
{OpenAI}.
\newblock Gpt-4v(ision).
\newblock \url{https://openai.com/research/gpt-4v-system-card}, 2023.
\newblock OpenAI.

\bibitem[Radford et~al.(2021)Radford, Kim, Hallacy, Ramesh, Goh, Agarwal, Sastry, Askell, Mishkin, Clark, Krueger, and Sutskever]{radford2021learningtransferablevisualmodels}
Alec Radford, Jong~Wook Kim, Chris Hallacy, Aditya Ramesh, Gabriel Goh, Sandhini Agarwal, Girish Sastry, Amanda Askell, Pamela Mishkin, Jack Clark, Gretchen Krueger, and Ilya Sutskever.
\newblock Learning transferable visual models from natural language supervision, 2021.
\newblock URL \url{https://arxiv.org/abs/2103.00020}.

\bibitem[Rao et~al.(2024)Rao, Choudhary, Deshpande, Satzoda, and Appalaraju]{rao2024ravenmultitaskretrievalaugmented}
Varun~Nagaraj Rao, Siddharth Choudhary, Aditya Deshpande, Ravi~Kumar Satzoda, and Srikar Appalaraju.
\newblock Raven: Multitask retrieval augmented vision-language learning, 2024.
\newblock URL \url{https://arxiv.org/abs/2406.19150}.

\bibitem[Singh et~al.(2019)Singh, Natarajan, Shah, Jiang, Chen, Batra, Parikh, and Rohrbach]{singh2019vqamodelsread}
Amanpreet Singh, Vivek Natarajan, Meet Shah, Yu~Jiang, Xinlei Chen, Dhruv Batra, Devi Parikh, and Marcus Rohrbach.
\newblock Towards vqa models that can read, 2019.
\newblock URL \url{https://arxiv.org/abs/1904.08920}.

\bibitem[Steiner et~al.(2024)Steiner, Pinto, Tschannen, Keysers, Wang, Bitton, Gritsenko, Minderer, Sherbondy, Long, Qin, Ingle, Bugliarello, Kazemzadeh, Mesnard, Alabdulmohsin, Beyer, and Zhai]{steiner2024paligemma2familyversatile}
Andreas Steiner, André~Susano Pinto, Michael Tschannen, Daniel Keysers, Xiao Wang, Yonatan Bitton, Alexey Gritsenko, Matthias Minderer, Anthony Sherbondy, Shangbang Long, Siyang Qin, Reeve Ingle, Emanuele Bugliarello, Sahar Kazemzadeh, Thomas Mesnard, Ibrahim Alabdulmohsin, Lucas Beyer, and Xiaohua Zhai.
\newblock Paligemma 2: A family of versatile vlms for transfer, 2024.
\newblock URL \url{https://arxiv.org/abs/2412.03555}.

\bibitem[Team et~al.(2024)Team, Riviere, Pathak, Sessa, Hardin, Bhupatiraju, Hussenot, Mesnard, Shahriari, Ramé, Ferret, Liu, Tafti, Friesen, Casbon, Ramos, Kumar, Lan, Jerome, Tsitsulin, Vieillard, Stanczyk, Girgin, Momchev, Hoffman, Thakoor, Grill, Neyshabur, Bachem, Walton, Severyn, Parrish, Ahmad, Hutchison, Abdagic, Carl, Shen, Brock, Coenen, Laforge, Paterson, Bastian, Piot, Wu, Royal, Chen, Kumar, Perry, Welty, Choquette-Choo, Sinopalnikov, Weinberger, Vijaykumar, Rogozińska, Herbison, Bandy, Wang, Noland, Moreira, Senter, Eltyshev, Visin, Rasskin, Wei, Cameron, Martins, Hashemi, Klimczak-Plucińska, Batra, Dhand, Nardini, Mein, Zhou, Svensson, Stanway, Chan, Zhou, Carrasqueira, Iljazi, Becker, Fernandez, van Amersfoort, Gordon, Lipschultz, Newlan, yeong Ji, Mohamed, Badola, Black, Millican, McDonell, Nguyen, Sodhia, Greene, Sjoesund, Usui, Sifre, Heuermann, Lago, McNealus, Soares, Kilpatrick, Dixon, Martins, Reid, Singh, Iverson, Görner, Velloso, Wirth, Davidow, Miller, Rahtz, Watson, Risdal,
  Kazemi, Moynihan, Zhang, Kahng, Park, Rahman, Khatwani, Dao, Bardoliwalla, Devanathan, Dumai, Chauhan, Wahltinez, Botarda, Barnes, Barham, Michel, Jin, Georgiev, Culliton, Kuppala, Comanescu, Merhej, Jana, Rokni, Agarwal, Mullins, Saadat, Carthy, Cogan, Perrin, Arnold, Krause, Dai, Garg, Sheth, Ronstrom, Chan, Jordan, Yu, Eccles, Hennigan, Kocisky, Doshi, Jain, Yadav, Meshram, Dharmadhikari, Barkley, Wei, Ye, Han, Kwon, Xu, Shen, Gong, Wei, Cotruta, Kirk, Rao, Giang, Peran, Warkentin, Collins, Barral, Ghahramani, Hadsell, Sculley, Banks, Dragan, Petrov, Vinyals, Dean, Hassabis, Kavukcuoglu, Farabet, Buchatskaya, Borgeaud, Fiedel, Joulin, Kenealy, Dadashi, and Andreev]{gemmateam2024gemma2improvingopen}
Gemma Team, Morgane Riviere, Shreya Pathak, Pier~Giuseppe Sessa, Cassidy Hardin, Surya Bhupatiraju, Léonard Hussenot, Thomas Mesnard, Bobak Shahriari, Alexandre Ramé, Johan Ferret, Peter Liu, Pouya Tafti, Abe Friesen, Michelle Casbon, Sabela Ramos, Ravin Kumar, Charline~Le Lan, Sammy Jerome, Anton Tsitsulin, Nino Vieillard, Piotr Stanczyk, Sertan Girgin, Nikola Momchev, Matt Hoffman, Shantanu Thakoor, Jean-Bastien Grill, Behnam Neyshabur, Olivier Bachem, Alanna Walton, Aliaksei Severyn, Alicia Parrish, Aliya Ahmad, Allen Hutchison, Alvin Abdagic, Amanda Carl, Amy Shen, Andy Brock, Andy Coenen, Anthony Laforge, Antonia Paterson, Ben Bastian, Bilal Piot, Bo~Wu, Brandon Royal, Charlie Chen, Chintu Kumar, Chris Perry, Chris Welty, Christopher~A. Choquette-Choo, Danila Sinopalnikov, David Weinberger, Dimple Vijaykumar, Dominika Rogozińska, Dustin Herbison, Elisa Bandy, Emma Wang, Eric Noland, Erica Moreira, Evan Senter, Evgenii Eltyshev, Francesco Visin, Gabriel Rasskin, Gary Wei, Glenn Cameron, Gus Martins,
  Hadi Hashemi, Hanna Klimczak-Plucińska, Harleen Batra, Harsh Dhand, Ivan Nardini, Jacinda Mein, Jack Zhou, James Svensson, Jeff Stanway, Jetha Chan, Jin~Peng Zhou, Joana Carrasqueira, Joana Iljazi, Jocelyn Becker, Joe Fernandez, Joost van Amersfoort, Josh Gordon, Josh Lipschultz, Josh Newlan, Ju~yeong Ji, Kareem Mohamed, Kartikeya Badola, Kat Black, Katie Millican, Keelin McDonell, Kelvin Nguyen, Kiranbir Sodhia, Kish Greene, Lars~Lowe Sjoesund, Lauren Usui, Laurent Sifre, Lena Heuermann, Leticia Lago, Lilly McNealus, Livio~Baldini Soares, Logan Kilpatrick, Lucas Dixon, Luciano Martins, Machel Reid, Manvinder Singh, Mark Iverson, Martin Görner, Mat Velloso, Mateo Wirth, Matt Davidow, Matt Miller, Matthew Rahtz, Matthew Watson, Meg Risdal, Mehran Kazemi, Michael Moynihan, Ming Zhang, Minsuk Kahng, Minwoo Park, Mofi Rahman, Mohit Khatwani, Natalie Dao, Nenshad Bardoliwalla, Nesh Devanathan, Neta Dumai, Nilay Chauhan, Oscar Wahltinez, Pankil Botarda, Parker Barnes, Paul Barham, Paul Michel, Pengchong Jin,
  Petko Georgiev, Phil Culliton, Pradeep Kuppala, Ramona Comanescu, Ramona Merhej, Reena Jana, Reza~Ardeshir Rokni, Rishabh Agarwal, Ryan Mullins, Samaneh Saadat, Sara~Mc Carthy, Sarah Cogan, Sarah Perrin, Sébastien M.~R. Arnold, Sebastian Krause, Shengyang Dai, Shruti Garg, Shruti Sheth, Sue Ronstrom, Susan Chan, Timothy Jordan, Ting Yu, Tom Eccles, Tom Hennigan, Tomas Kocisky, Tulsee Doshi, Vihan Jain, Vikas Yadav, Vilobh Meshram, Vishal Dharmadhikari, Warren Barkley, Wei Wei, Wenming Ye, Woohyun Han, Woosuk Kwon, Xiang Xu, Zhe Shen, Zhitao Gong, Zichuan Wei, Victor Cotruta, Phoebe Kirk, Anand Rao, Minh Giang, Ludovic Peran, Tris Warkentin, Eli Collins, Joelle Barral, Zoubin Ghahramani, Raia Hadsell, D.~Sculley, Jeanine Banks, Anca Dragan, Slav Petrov, Oriol Vinyals, Jeff Dean, Demis Hassabis, Koray Kavukcuoglu, Clement Farabet, Elena Buchatskaya, Sebastian Borgeaud, Noah Fiedel, Armand Joulin, Kathleen Kenealy, Robert Dadashi, and Alek Andreev.
\newblock Gemma 2: Improving open language models at a practical size, 2024.
\newblock URL \url{https://arxiv.org/abs/2408.00118}.

\bibitem[Wang et~al.(2022)Wang, Zhou, Zeng, and Zhang]{wang2022efficientvlm}
Tiannan Wang, Wangchunshu Zhou, Yan Zeng, and Xinsong Zhang.
\newblock Efficientvlm: Fast and accurate vision-language models via knowledge distillation and modal-adaptive pruning.
\newblock \emph{arXiv preprint arXiv:2210.07795}, 2022.

\bibitem[Yang et~al.(2024)Yang, Yang, Hui, Zheng, Yu, Zhou, Li, Li, Liu, Huang, Dong, Wei, Lin, Tang, Wang, Yang, Tu, Zhang, Ma, Yang, Xu, Zhou, Bai, He, Lin, Dang, Lu, Chen, Yang, Li, Xue, Ni, Zhang, Wang, Peng, Men, Gao, Lin, Wang, Bai, Tan, Zhu, Li, Liu, Ge, Deng, Zhou, Ren, Zhang, Wei, Ren, Liu, Fan, Yao, Zhang, Wan, Chu, Liu, Cui, Zhang, Guo, and Fan]{yang2024qwen2technicalreport}
An~Yang, Baosong Yang, Binyuan Hui, Bo~Zheng, Bowen Yu, Chang Zhou, Chengpeng Li, Chengyuan Li, Dayiheng Liu, Fei Huang, Guanting Dong, Haoran Wei, Huan Lin, Jialong Tang, Jialin Wang, Jian Yang, Jianhong Tu, Jianwei Zhang, Jianxin Ma, Jianxin Yang, Jin Xu, Jingren Zhou, Jinze Bai, Jinzheng He, Junyang Lin, Kai Dang, Keming Lu, Keqin Chen, Kexin Yang, Mei Li, Mingfeng Xue, Na~Ni, Pei Zhang, Peng Wang, Ru~Peng, Rui Men, Ruize Gao, Runji Lin, Shijie Wang, Shuai Bai, Sinan Tan, Tianhang Zhu, Tianhao Li, Tianyu Liu, Wenbin Ge, Xiaodong Deng, Xiaohuan Zhou, Xingzhang Ren, Xinyu Zhang, Xipin Wei, Xuancheng Ren, Xuejing Liu, Yang Fan, Yang Yao, Yichang Zhang, Yu~Wan, Yunfei Chu, Yuqiong Liu, Zeyu Cui, Zhenru Zhang, Zhifang Guo, and Zhihao Fan.
\newblock Qwen2 technical report, 2024.
\newblock URL \url{https://arxiv.org/abs/2407.10671}.

\bibitem[Yang et~al.(2023{\natexlab{a}})Yang, Li, Wang, Lin, Azarnasab, Ahmed, Liu, Liu, Zeng, and Wang]{yang2023mmreactpromptingchatgptmultimodal}
Zhengyuan Yang, Linjie Li, Jianfeng Wang, Kevin Lin, Ehsan Azarnasab, Faisal Ahmed, Zicheng Liu, Ce~Liu, Michael Zeng, and Lijuan Wang.
\newblock Mm-react: Prompting chatgpt for multimodal reasoning and action, 2023{\natexlab{a}}.
\newblock URL \url{https://arxiv.org/abs/2303.11381}.

\bibitem[Yang et~al.(2023{\natexlab{b}})Yang, Ping, Liu, Korthikanti, Nie, Huang, Fan, Yu, Lan, Li, Liu, Zhu, Shoeybi, Catanzaro, Xiao, and Anandkumar]{yang2023revilmretrievalaugmentedvisuallanguage}
Zhuolin Yang, Wei Ping, Zihan Liu, Vijay Korthikanti, Weili Nie, De-An Huang, Linxi Fan, Zhiding Yu, Shiyi Lan, Bo~Li, Ming-Yu Liu, Yuke Zhu, Mohammad Shoeybi, Bryan Catanzaro, Chaowei Xiao, and Anima Anandkumar.
\newblock Re-vilm: Retrieval-augmented visual language model for zero and few-shot image captioning, 2023{\natexlab{b}}.
\newblock URL \url{https://arxiv.org/abs/2302.04858}.

\bibitem[Yao et~al.(2024)Yao, Yu, Zhang, Wang, Cui, Zhu, Cai, Li, Zhao, He, Chen, Zhou, Zou, Zhang, Hu, Zheng, Zhou, Cai, Han, Zeng, Li, Liu, and Sun]{yao2024minicpmvgpt4vlevelmllm}
Yuan Yao, Tianyu Yu, Ao~Zhang, Chongyi Wang, Junbo Cui, Hongji Zhu, Tianchi Cai, Haoyu Li, Weilin Zhao, Zhihui He, Qianyu Chen, Huarong Zhou, Zhensheng Zou, Haoye Zhang, Shengding Hu, Zhi Zheng, Jie Zhou, Jie Cai, Xu~Han, Guoyang Zeng, Dahai Li, Zhiyuan Liu, and Maosong Sun.
\newblock Minicpm-v: A gpt-4v level mllm on your phone, 2024.
\newblock URL \url{https://arxiv.org/abs/2408.01800}.

\bibitem[Yu et~al.(2024)Yu, Yang, Li, Wang, Lin, Liu, Wang, and Wang]{yu2024mmvetevaluatinglargemultimodal}
Weihao Yu, Zhengyuan Yang, Linjie Li, Jianfeng Wang, Kevin Lin, Zicheng Liu, Xinchao Wang, and Lijuan Wang.
\newblock Mm-vet: Evaluating large multimodal models for integrated capabilities, 2024.
\newblock URL \url{https://arxiv.org/abs/2308.02490}.

\bibitem[Yuan et~al.(2024)Yuan, Li, Huang, Ye, and Sun]{yuan2024tinygptvefficientmultimodallarge}
Zhengqing Yuan, Zhaoxu Li, Weiran Huang, Yanfang Ye, and Lichao Sun.
\newblock Tinygpt-v: Efficient multimodal large language model via small backbones, 2024.
\newblock URL \url{https://arxiv.org/abs/2312.16862}.

\bibitem[Yue et~al.(2024)Yue, Ni, Zhang, Zheng, Liu, Zhang, Stevens, Jiang, Ren, Sun, Wei, Yu, Yuan, Sun, Yin, Zheng, Yang, Liu, Huang, Sun, Su, and Chen]{yue2024mmmumassivemultidisciplinemultimodal}
Xiang Yue, Yuansheng Ni, Kai Zhang, Tianyu Zheng, Ruoqi Liu, Ge~Zhang, Samuel Stevens, Dongfu Jiang, Weiming Ren, Yuxuan Sun, Cong Wei, Botao Yu, Ruibin Yuan, Renliang Sun, Ming Yin, Boyuan Zheng, Zhenzhu Yang, Yibo Liu, Wenhao Huang, Huan Sun, Yu~Su, and Wenhu Chen.
\newblock Mmmu: A massive multi-discipline multimodal understanding and reasoning benchmark for expert agi, 2024.
\newblock URL \url{https://arxiv.org/abs/2311.16502}.

\bibitem[Zhai et~al.(2023)Zhai, Mustafa, Kolesnikov, and Beyer]{zhai2023sigmoidlosslanguageimage}
Xiaohua Zhai, Basil Mustafa, Alexander Kolesnikov, and Lucas Beyer.
\newblock Sigmoid loss for language image pre-training, 2023.
\newblock URL \url{https://arxiv.org/abs/2303.15343}.

\bibitem[Zhang et~al.(2024)Zhang, Zeng, Wang, and Lu]{zhang2024tinyllamaopensourcesmalllanguage}
Peiyuan Zhang, Guangtao Zeng, Tianduo Wang, and Wei Lu.
\newblock Tinyllama: An open-source small language model, 2024.
\newblock URL \url{https://arxiv.org/abs/2401.02385}.

\bibitem[Zhou et~al.(2024)Zhou, Hu, Weng, Jia, Luo, Liu, Wu, and Huang]{zhou2024tinyllavaframeworksmallscalelarge}
Baichuan Zhou, Ying Hu, Xi~Weng, Junlong Jia, Jie Luo, Xien Liu, Ji~Wu, and Lei Huang.
\newblock Tinyllava: A framework of small-scale large multimodal models, 2024.
\newblock URL \url{https://arxiv.org/abs/2402.14289}.

\bibitem[Zhu et~al.(2023)Zhu, Chen, Shen, Li, and Elhoseiny]{zhu2023minigpt4enhancingvisionlanguageunderstanding}
Deyao Zhu, Jun Chen, Xiaoqian Shen, Xiang Li, and Mohamed Elhoseiny.
\newblock Minigpt-4: Enhancing vision-language understanding with advanced large language models, 2023.
\newblock URL \url{https://arxiv.org/abs/2304.10592}.

\bibitem[Zhu et~al.(2024)Zhu, Feng, Du, Gu, Yu, Wang, Chen, Chu, Chen, and Qin]{zhu2024informationbottleneckperspectiveeffective}
Kun Zhu, Xiaocheng Feng, Xiyuan Du, Yuxuan Gu, Weijiang Yu, Haotian Wang, Qianglong Chen, Zheng Chu, Jingchang Chen, and Bing Qin.
\newblock An information bottleneck perspective for effective noise filtering on retrieval-augmented generation, 2024.
\newblock URL \url{https://arxiv.org/abs/2406.01549}.

\end{thebibliography}
\newpage 

\appendix
\section{Detailed Analysis on Enhancing Effective Mutual Information via RAG}
\label{sec:appendix_theory}
As discussed, Lightweight, frozen LLMs (\(\theta_{\text{LLM,small}}\)) often exhibit performance limitations due to a substantial irreducible error \(\bar{\epsilon}_{\theta_{\text{LLM}}}\). To mitigate this, we introduce TinyAlign (Fig.~\ref{fig:tinyalign_architecture}), a Retrieval-Augmented Generation (RAG)-enhanced connector architecture. This approach boosts \textit{effective mutual information} (\(\Ieff\)) by supplying strategically compressed, highly relevant contextual cues, thereby reducing the intrinsic processing demands on the frozen LLM.

A standard VLM processes a visual input \(X_V\) via a ViT (\(\theta_{\text{ViT}}\)) to obtain \(Z_V\), which a primary connector (\(\theta_C^*\)) maps to \(H_V\). Instruction embeddings \(H_I\) are also generated. TinyAlign augments this by:
1) retrieving \(k\) relevant, pre-compressed embeddings \(E_R = \{E_{R_j}\}_{j=1}^k\) from a memory bank \(\mathcal{M}\) (Sec.~\ref{subsec:memory_design});
2) employing a trainable RAG connector (\(\theta_{RC}^*\)) to transform \(E_R\) into supplementary representations \(H_R\); and
3) presenting a composite input \(H_{\text{in}}' = [H_V, H_R, H_I]\) to the frozen LLM. The trainable parameters are thus \(\Theta_C^* = \{\theta_C^*, \theta_{RC}^*\}\).
We posit that incorporating \(E_R\)—forming an augmented context \(X' = (X_V, X_I, E_R)\)—enhances \(\Ieff(X'; L | \theta_{\text{LLM}}, \theta_{\text{ViT}})\). The change, \(\Delta \Ieff\), is:
\begin{align}
\Delta \Ieff &= [I(X'; L) - \bar{\epsilon}_{\theta_{\text{LLM}}}(X')] - [I(X_V, X_I; L) - \bar{\epsilon}_{\theta_{\text{LLM}}}(X_V, X_I)] \nonumber \\
&= \underbrace{I(E_R; L | X_V, X_I)}_{\Delta I_{\text{true}}} + \underbrace{(\bar{\epsilon}_{\theta_{\text{LLM}}}(X_V, X_I) - \bar{\epsilon}_{\theta_{\text{LLM}}}(X'))}_{\Delta \bar{\epsilon}_{\theta_{\text{LLM}}}} 
\end{align}
The first term, \(\Delta I_{\text{true}}\), is positive because \(E_R\), derived from pertinent captions (i.e., clues for the target \(L\)), provides novel information about \(L\) conditioned on \(X_V\) and \(X_I\).
The second term, \(\Delta \bar{\epsilon}_{\theta_{\text{LLM}}}\), signifies a positive reduction in the LLM's irreducible error. The RAG connector \(\theta_{RC}^*\) transforms \(E_R\) into \(H_R\), providing what we term 'concise, LLM-assimilable contextual hints.' These hints—originating from pre-compressed image-text pairs (potent clues processed by \(\theta_P\))—present information in a format more attuned to the LLM's textual processing strengths than deciphering complex visual semantics solely from \(H_V\). This enhanced 'input friendliness' of the augmented input \(H_{\text{in}}'\) enables the fixed-capacity LLM to approximate the target distribution \(\Ptrue(L|X')\) with greater fidelity than it could from \((X_V, X_I)\) alone, a phenomenon corroborated by faster convergence (Fig.~\ref{fig:combined_loss_and_umap}(a)). This improved fidelity directly translates into fewer fundamental misinterpretations, thereby lowering the irreducible error: \(\bar{\epsilon}_{\theta_{\text{LLM}}}(X') < \bar{\epsilon}_{\theta_{\text{LLM}}}(X_V, X_I)\). The trainable RAG connector \(\theta_{RC}^*\) is optimized to generate \(H_R\) that maximizes this error reduction by effectively complementing \(H_V\) and \(H_I\), thus alleviating the LLM's cognitive burden.

This reduction in irreducible error is especially impactful for lightweight VLMs, whose inherent capacity constraints often yield a higher baseline \(\bar{\epsilon}_{\theta_{\text{LLM}}}(X_V, X_I)\). TinyAlign's architecture—emphasizing LLM-independent, Perceiver-based pre-compression for the memory bank and an efficient RAG connector—is engineered to maximize this \(\Delta \bar{\epsilon}_{\theta_{\text{LLM}}}\). It functions as a 'cognitive scaffold,' substantially lowering the reasoning threshold for these less capable LLMs—a benefit less critical for LLMs endowed with robust intrinsic reasoning.
Consequently, when \(\Delta \Ieff > 0\)—driven by both \(\Delta I_{\text{true}} > 0\) and a substantial \(\Delta \bar{\epsilon}_{\theta_{\text{LLM}}} > 0\) for lightweight LLMs—the system's minimum achievable CE loss is reduced: \(\min_{\theta_C^*} \LCE(\text{RAG-enhanced}) < \min_{\theta_C^*} \LCE(\text{standard})\). This framework demonstrates that RAG-enhanced connectors, by strategically mitigating the processing burden and inherent error of a frozen (especially lightweight) LLM using pre-compressed multimodal cues, can elevate VLM performance. This effectively increases the information the LLM can leverage despite its fixed capacity.
\section{Hyperparameter Summary}
\label{sec:appendix_hyperparameters}
This subsection provides a comprehensive overview of the critical hyperparameters employed throughout our experimental phases, encompassing both pre-training and fine-tuning stages for our proposed models, as well as the specific configuration for the Perceiver model utilized in TinyAlign's memory bank construction. Table~\ref{tab:hyperparameters} delineates the comparative settings for pre-training and fine-tuning, covering aspects such as batch sizes, learning rates, and optimization strategies. Complementing this, Table~\ref{tab:perceiver_config} itemizes the architectural hyperparameters of the Perceiver model, detailing its latent space dimensions, attention mechanisms, and input specifications. All experiments were conducted on a system equipped with eight NVIDIA RTX 4090 GPUs, each with 48GB of VRAM, ensuring robust and scalable computational support.

\begin{table}[htbp]
\centering
\caption{Key hyperparameters for pre-training and fine-tuning.}
\label{tab:hyperparameters}
\begin{tabular}{@{}lcc@{}}
\toprule
Hyperparameter                 & Pre-training Value & Fine-tuning Value \\
\midrule
Global Batch Size              & 256                & 128               \\
Per-device Batch Size (script) & 16                 & 12                \\
Gradient Accumulation (script) & 1                  & 2                 \\
Learning Rate                  & 1e-3               & 5e-8              \\
LR Scheduler                   & Cosine             & Cosine            \\
Warmup Ratio                   & 0.03               & 0.03              \\
Precision                      & FP16               & FP16              \\
Optimizer                      & AdamW              & AdamW  \\
LLM Tuning                     & Frozen             & Full              \\
Vision Tower Tuning            & Frozen             & Frozen            \\
Connector Tuning               & Full               & Full              \\
\bottomrule
\end{tabular}
\end{table}

\begin{table}[htbp]
\centering
\caption{PerceiverConfig Hyperparameters}
\label{tab:perceiver_config}
\begin{tabular}{@{}lll@{}}
\toprule
\textbf{Parameter} & \textbf{Value}  \\
\midrule
\texttt{num\_latents} & 32  \\
\texttt{d\_latents} & 96 \\
\texttt{d\_model} & 128  \\
\texttt{num\_self\_attends\_per\_block} & 8  \\
\texttt{num\_blocks} & 1 \\
\texttt{num\_self\_attention\_heads} & 8 \\
\texttt{num\_cross\_attention\_heads} & 8 \\
\texttt{qk\_channels} & 96  \\
\texttt{v\_channels} & 96 \\
\texttt{image\_size} & 384 \\
\texttt{vocab\_size} & 30522  \\
\texttt{max\_position\_embeddings} & 512 \\
\bottomrule
\end{tabular}
\end{table}

\section{UMAP Visualization of Feature Embeddings}
\label{sec:appendix_umap}
To elucidate the latent structure and interrelations between image-derived and text-derived features, we employed Uniform Manifold Approximation and Projection (UMAP) for dimensionality reduction and visualization.

\textbf{UMAP Fundamentals:}
UMAP is a non-linear dimensionality reduction algorithm adept at preserving both local and global structures of high-dimensional data within a lower-dimensional embedding. It achieves this by first constructing a high-dimensional graph representation of the data, where edge weights denote the likelihood of connectivity between points. Subsequently, it optimizes a low-dimensional graph to maximize structural similarity to its high-dimensional counterpart. This methodology is rooted in Riemannian geometry and algebraic topology, aiming to model the data's underlying manifold.

\textbf{Visualization Methodology:}
Our UMAP-based visualization procedure comprised the following steps:

\begin{enumerate}
\item \textbf{Input Data Preparation}:
The UMAP process utilized two sets of high-dimensional feature vectors:
\begin{itemize}
\item \textbf{Connector Features}: Vectors derived from images post-processing by the vision tower and the model's connector module. These represent visual information conditioned for the language model.
\item \textbf{LLM Input Embeddings}: Corresponding vectors representing embeddings of textual inputs (e.g., questions, prompts) from the language model's input embedding layer.
\end{itemize}
These two feature sets were concatenated. Origin labels (connector feature or LLM embedding) were retained for subsequent differentiated visualization.
\item \textbf{UMAP Dimensionality Reduction}:
The UMAP algorithm was applied to the concatenated high-dimensional dataset, configured to reduce feature dimensionality to two. This 2D representation facilitates direct scatter plot visualization. The algorithm learns a mapping that optimally preserves the topological structure of the original feature space in this lower dimension.
\item \textbf{Output and Visualization}:
The UMAP process yielded:
\begin{itemize}
    \item \textbf{2D Coordinates}: A two-dimensional coordinate (x, y) for each input feature vector (both connector and LLM embedding).
    \item \textbf{Scatter Plot Visualization}: These 2D coordinates were used to generate a scatter plot. Points corresponding to connector features and LLM input embeddings were rendered in distinct colors. This visualization enables qualitative assessment of:
    \begin{itemize}
        \item The distinctness or overlap between the two feature spaces.
        \item The presence of intra-feature-type clusters.
        \item The overall geometric relationship between the model's latent representations of visual and textual information.
    \end{itemize}
\end{itemize}

\end{enumerate}

\section{FLOPs Analysis}
\label{sec:appendix_flops_analysis}

This section presents a comparative analysis of the Floating Point Operations (FLOPs) for models based on the Phi-2 architecture, focusing on the computational demands during different stages. As detailed in Table~\ref{tab:flops_analysis_table}, we evaluate both our proposed TinyAlign approach and a baseline model across pre-training and fine-tuning phases. The analysis indicates that while the introduction of RAG tokens and an additional RAG connector in TinyAlign incurs a marginal increase in computational operations, the overall impact on inference efficiency and training throughput remains negligible, underscoring the computational viability of our method.

\begin{table}[htbp]
\centering
\caption{FLOPs analysis for Phi-2 based models. While the introduction of RAG tokens and an additional RAG connector incurs a marginal increase in computational operations, the impact on overall inference efficiency is negligible.}
\label{tab:flops_analysis_table}
\begin{tabular}{lcc}
\toprule
Model Stage             & Train Steps/Second & FLOPs   \\
\midrule
TinyAlign-finetune    & 0.075              & 6.15e18 \\
TinyAlign-pretrain    & 0.112              & 2.45e14 \\
Baseline-finetune      & 0.083              & 6.12e18 \\
Baseline-pretrain      & 0.173              & 2.05e14 \\
\bottomrule
\end{tabular}
\end{table}

\section{Detailed Data Efficiency Analysis}
\label{sec:appendix_detailed_data_efficiency}

\begin{figure}[htbp]
  \centering
  \includegraphics[width=1.0\linewidth]{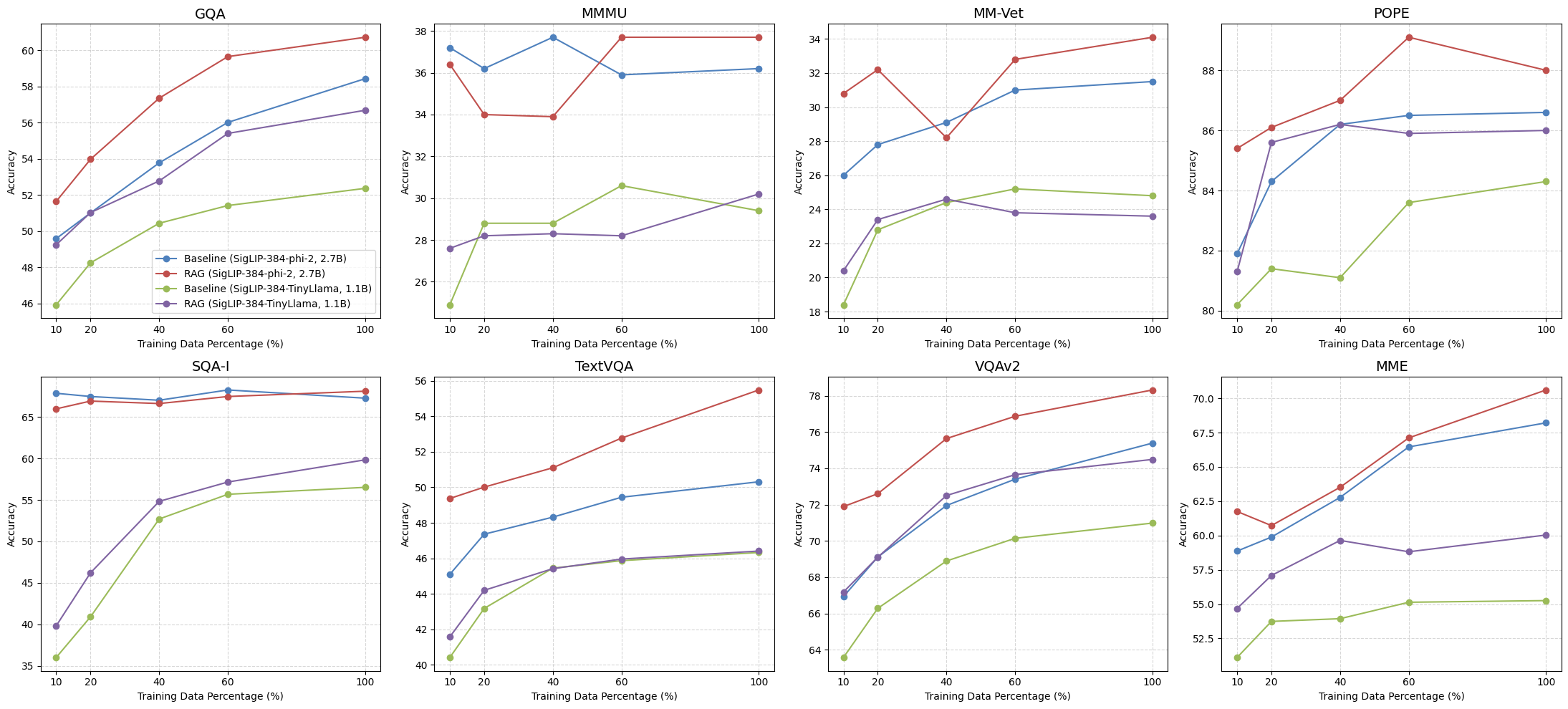} 
  \caption{Detailed data efficiency analysis across individual benchmarks. Performance of TinyAlign-enhanced models (variants of Phi-2 and TinyLLaMA) is compared against their respective baselines at varying percentages of instruction tuning data.}
  \label{fig:bech_appendix} 
\end{figure}

TinyAlign models exhibit remarkable data efficiency during instruction tuning, as illustrated comparatively in Fig.~\ref{fig:data_efficiency_main} (main text) and further detailed per benchmark in Fig.~\ref{fig:bech_appendix}. Our comprehensive analysis indicates that TinyAlign-enhanced models, specifically LightweightLLM-Phi-2 and TinyLLaMA, can match the weighted average accuracy of baseline models trained with 100\% of the instruction tuning data while using only 40\% of this data. A granular examination of individual benchmarks further substantiates TinyAlign's effectiveness in learning from limited labeled data and enhancing critical VLM capabilities.

\paragraph{Enhanced Visual Reasoning with Greater Efficiency.}
On visual reasoning benchmarks, TinyAlign yields substantial data efficiency gains. For \textbf{GQA}, TinyAlign (SigLIP-384-phi-2, 2.7B) with 60\% data (59.7 accuracy) surpasses the baseline with 100\% data (58.4 accuracy). Similarly, TinyAlign (SigLIP-384-TinyLlama, 1.1B) with just 40\% data (52.78 accuracy) outperforms its baseline counterpart using 100\% data (52.37 accuracy). On \textbf{VQAv2}, TinyAlign (SigLIP-384-phi-2, 2.7B) at 40\% data (75.64 accuracy) matches the baseline's full-data performance (75.39 accuracy), while TinyAlign (SigLIP-384-TinyLlama, 1.1B) with 40\% data (72.50 accuracy) significantly exceeds its baseline with 100\% data (70.98 accuracy). For \textbf{SQA-I}, the TinyAlign-enhanced TinyLlama model demonstrates a particularly notable improvement in its performance trajectory with increasing data efficiency, compared to a relatively stagnant baseline.

\paragraph{Improved Scene Text Understanding.}
TinyAlign's advantages extend to scene text understanding. On \textbf{TextVQA}, TinyAlign (SigLIP-384-phi-2, 2.7B) utilizing only 40\% of the training data (51.10 accuracy) achieves higher accuracy than the baseline model trained with 100\% data (50.31 accuracy), suggesting an improved ability to integrate visual and textual cues from sparser signals.

\paragraph{Effective Suppression of Model Hallucinations.}
TinyAlign significantly enhances model factuality by reducing hallucinations, as evidenced on the \textbf{POPE} benchmark. TinyAlign models consistently achieve higher accuracy (implying fewer hallucinations) across various data proportions. For instance, TinyAlign (SigLIP-384-TinyLlama, 1.1B) with only 20\% data (85.6 accuracy) surpasses the baseline TinyLlama trained with 100\% data (84.3 accuracy). TinyAlign (SigLIP-384-phi-2, 2.7B) exhibits similar data efficiency, reaching the baseline's peak performance with only 40\% of the data.

\paragraph{Broad Applicability and Efficacy on Lightweight Models.}
On comprehensive benchmarks such as \textbf{MME}, TinyAlign generally provides consistent performance improvements for both Phi-2 and TinyLlama backbones. For instance, TinyAlign-enhanced models show a clear advantage over baselines across most data regimes on MME. On \textbf{MM-Vet}, data efficiency is also prominent: TinyAlign (SigLIP-384-phi-2, 2.7B) with 20\% data (32.2 accuracy) outperforms the baseline with 100\% data (31.5 accuracy), while TinyAlign (SigLIP-384-TinyLlama, 1.1B) with 40\% data (24.6 accuracy) performs comparably to its baseline with 100\% data (24.8 accuracy).

\paragraph{Addressing Challenges in Complex Tasks.}
While TinyAlign demonstrates broad advantages, its impact can vary with task complexity. On the highly intricate \textbf{MMMU} benchmark, performance gains from TinyAlign are less consistent across data proportions compared to other benchmarks. Furthermore, on \textbf{MM-Vet}, the baseline TinyLlama model slightly outperforms its TinyAlign counterpart when both are trained with 100\% data. These instances suggest avenues for future research, such as optimizing TinyAlign for extremely complex tasks or specific model-data interactions.

To comprehensively evaluate the model's overall performance across multiple distinct benchmark datasets, we employed \textbf{weighted average accuracy}. This method assigns a corresponding weight based on the sample size (or representative importance) \(C_i\) of each test set. For the accuracies \(Acc_i\) obtained for each test set under specific conditions (e.g., a certain training data percentage), the weighted average accuracy \(A_{\text{weighted}}\) is calculated as follows:
\begin{align}
\label{eq:acc}
A_{\text{weighted}} = \frac{\sum (Acc_i \times C_i)}{\sum C_i}
\end{align}
This approach allows us to more accurately reflect the model's combined performance on datasets of varying scales and characteristics, avoiding potential biases from simple averaging.

\section{Detailed Ablation Studies}
\label{sec:appendix_detailed_Ablation_Studies}
We conduct several ablation studies to validate key design choices within the TinyAlign framework. These studies focus on knowledge base (KB) size, vision encoder alignment for RAG key generation, and the number of retrieved documents (Top-K). All ablations use the Phi-2 (2.7B) model as the LLM backbone unless otherwise specified.

\paragraph{Knowledge Base Size.}
We evaluate KB sizes of 100k, 300k, and 500k entries, with retrieval fixed at top-10 documents (Table~\ref{tab:ablation_kb_size_appendix}). The 100k KB shows comparable, and in some cases superior, performance to larger KBs while offering greater efficiency. This finding suggests that a moderately sized, high-quality KB is sufficient and eliminates the need for excessively large memory banks.

\begin{table}[htbp]
\centering
\caption{Ablation on knowledge base (KB) size for Phi-2 (2.7B) with top-10 retrieval. Performance reported on various benchmarks.}
\label{tab:ablation_kb_size_appendix} 
\resizebox{\textwidth}{!}{%
\begin{tabular}{@{}lcccccccc@{}}
\toprule
KB Size & GQA & MMMU & MM-Vet & POPE & SQA-I & TextVQA & VQAv2 & MME \\
\midrule
100k    & 60.73 & 37.7 & 34.1 & 0.880 & 68.12 & 55.48 & 78.32 & 1412.1 \\
300k    & 61.27 & 36.8 & 31.7 & 0.871 & 70.25 & 55.42 & 78.58 & 1407.34 \\
500k    & 61.38 & 36.0 & 31.6 & 0.875 & 68.91 & 56.43 & 78.57 & 1401.06 \\
\bottomrule
\end{tabular}%
}
\end{table}

\paragraph{Vision Encoder Alignment for RAG Keys.}
We investigate the importance of aligning the vision encoder used for VLM pre-training with the encoder used for generating RAG keys (Table~\ref{tab:ablation_vit_alignment_appendix}). Using \texttt{stablelm-2-zephyr-1.6b} as the VLM, we compare a matched configuration (VLM pre-trained with SigLIP features, RAG keys from SigLIP) against a baseline VLM (pre-trained with CLIP features, no RAG) and a mismatched RAG configuration (VLM pre-trained with CLIP features, RAG keys from SigLIP). The results show significant performance degradation, including a "collapse" on benchmarks like VQAv2, when encoders are mismatched. This highlights the critical importance of feature space consistency between the VLM's learned representations and the RAG system's retrieval mechanism. The "TinyAlign (Matched)" configuration, representing our proposed approach, consistently performs on par with or better than the CLIP VLM baseline, demonstrating the advantages of properly aligned RAG systems.

\begin{table}[htbp]
\centering
\caption{Ablation on vision encoder alignment. VLM: \texttt{stablelm-2-zephyr-1.6b}. KB: 100k SigLIP-keyed. "Matched": VLM (SigLIP features) + RAG (SigLIP keys). "Baseline": VLM (CLIP features), no RAG. "Mismatched": VLM (CLIP features) + RAG (SigLIP keys).}
\label{tab:ablation_vit_alignment_appendix} 
\resizebox{\textwidth}{!}{%
\begin{tabular}{@{}lccc@{}}
\toprule
Benchmark & TinyAlign (Matched: SigLIP VLM + SigLIP RAG) & Baseline (CLIP VLM) & TinyAlign (Mismatched: CLIP VLM + SigLIP RAG) \\
\midrule
GQA       & 51.28  & 52.73  & 40.96  \\
MMMU      & 0.304  & 0.313  & 0.300  \\
MM-Vet    & 28.4   & 28.4   & 24.6   \\
POPE      & 0.847  & 0.836  & 0.783  \\
SQA-I     & 62.32  & 62.47  & 56.17  \\
TextVQA   & 49.39  & 46.75  & 36.71  \\
VQAv2     & 70.12  & 69.82  & 29.62  \\ 
MME       & 1218.40& 1206.49& 1054.78\\
\bottomrule
\end{tabular}%
}
\end{table}
\begin{table}[htbp]
\centering
\caption{Ablation on top-K retrieval for Phi-2 (2.7B) with a 100k knowledge base.}
\label{tab:ablation_top_k_phi_appendix} 
\resizebox{\textwidth}{!}{%
\begin{tabular}{@{}lcccccccc@{}}
\toprule
Top-K & GQA & MMMU & MM-Vet & POPE & SQA-I & TextVQA & VQAv2 & MME \\
\midrule
Top-1   & 60.6  & 36.0 & 33.2 & 0.874 & 70.35 & 55.66 & 78.21 & 1411.12 \\
Top-5   & 61.27 & 38.1 & 33.6 & 0.874 & 70.00 & 55.69 & 78.48 & 1415.22 \\ 
Top-10  & 60.73 & 37.7 & 34.1 & 0.880 & 68.12 & 55.48 & 78.32 & 1412.1 \\
\bottomrule
\end{tabular}%
}
\end{table}
\paragraph{Top-K Retrieval.}
Using the 100k KB, we ablate the number of retrieved documents (Top-K) for Phi-2, comparing top-1, top-5, and top-10 retrieval (Table~\ref{tab:ablation_top_k_phi_appendix}). Retrieving the top-5 documents achieves the best overall balance of performance across benchmarks, providing richer contextual signals than top-1 while avoiding the potential noise or diminishing returns observed with top-10 in some cases.

\end{document}